\documentclass[acmsmall]{acmart}

\usepackage{multirow}
\usepackage{graphicx}
\usepackage{subfig}
\graphicspath{ {images/} }
\usepackage[english]{babel}
\usepackage[utf8]{inputenc}
\usepackage{hyperref}

\usepackage{collcell}
\usepackage{hhline}
\usepackage{pgf}
\usepackage{multirow}
\usepackage{diagbox}
\usepackage{tikz}
\usetikzlibrary{calc}
\usepackage {color, graphicx}
\usetikzlibrary{decorations.pathmorphing}
\usetikzlibrary {patterns,shapes,shadows,decorations.pathreplacing}
\usepackage{amssymb}

\newcommand{\etal}{\mbox{\emph{et al.}}}

\hypersetup{portuguese,
    colorlinks=true,  
    linkcolor=blue,  
    citecolor=blue, 
    filecolor=blue,   
    urlcolor=blue,
    breaklinks=true,
}
   
\setcopyright{rightsretained}

\acmDOI{XX.XXX/XXX_X}

\acmISBN{XXX-XXXX-XX-XXX/XX/XX}

\acmJournal{TOIS}
\acmVolume{0}
\acmNumber{0}
\acmArticle{0}
\acmYear{2019}
\acmMonth{0}
\copyrightyear{2019}
\acmPrice{0}

\begin{document}
\sloppy

\clubpenalty=10000 
\widowpenalty = 10000

\title[Sentiment Analysis of Urban Outdoor Images by Using Semantic and Deep Features]{OutdoorSent: Sentiment Analysis of Urban Outdoor Images by Using Semantic and Deep Features}

\author{Wyverson Bonasoli de Oliveira}
 \affiliation{
   \institution{Universidade Tecnol\'{o}gica Federal do Paran\'{a} - UTFPR}
   \city{Curitiba} 
   \state{Paran\'{a}, Brazil}
 }
 \author{Leyza Baldo Dorini}
 \affiliation{
   \institution{Universidade Tecnol\'{o}gica Federal do Paran\'{a} - UTFPR}
   \city{Curitiba} 
   \state{Paran\'{a}, Brazil}
 }
 \author{Rodrigo Minetto}
 \affiliation{
   \institution{Universidade Tecnol\'{o}gica Federal do Paran\'{a} - UTFPR}
   \city{Curitiba} 
   \state{Paran\'{a}, Brazil}
 }
 \author{Thiago H. Silva}
 \authornote{Thiago initiated this study while at UTFPR.}
 \affiliation{
   \institution{University of Toronto}
   \city{Toronto} 
   \state{Canada}
 }
 \email{thiagoh@utfpr.edu.br}
  

\begin{abstract}
Opinion mining in outdoor images posted by users during different activities can provide valuable information to better understand urban areas. In this regard, we propose a framework to classify the sentiment of outdoor images shared by users on social networks. We compare the performance of state-of-the-art ConvNet architectures, and one specifically designed for sentiment analysis. We also evaluate how the merging of deep features and semantic information derived from the scene attributes can improve classification and cross-dataset generalization performance. The evaluation explores a novel dataset, namely OutdoorSent, and other datasets publicly available. We observe that the incorporation of knowledge about semantic attributes improves the accuracy of all ConvNet architectures studied. Besides, we found that exploring only images related to the context of the study, outdoor in our case, is recommended, i.e., indoor images were not significantly helpful. Furthermore, we demonstrated the applicability of our results in the city of Chicago, USA, showing that they can help to improve the knowledge of subjective characteristics of different areas of the city. For instance, particular areas of the city tend to concentrate more images of a specific class of sentiment, which are also correlated with median income, opening up opportunities in different fields.
\end{abstract}

%
%



\setcopyright{acmcopyright}
\acmJournal{TOIS}
\acmYear{2020} \acmVolume{1} \acmNumber{1} \acmArticle{1} \acmMonth{1} \acmPrice{15.00}\acmDOI{}

\maketitle

\section{Introduction}

Sentiment analysis based on images has an inherent subjectivity since it involves the visual recognition of objects, scenes, actions, and events.
Therefore, the algorithms need to be robust, make use of different
types of features, and have a generalization capability to cover different
domains. Even so, the problem is still challenging because distinct people may have different opinions (in terms of sentiment polarity) about the same image.

Nowadays, a considerable amount of people share their experiences and opinions on the most diverse subjects on online social networks. This  generates a vast amount of data, and their proper analysis plays an essential role in several segments, ranging from  prediction of spatial events ~\cite{Zhao:2016:OSE:3014433.2997642}, and rumor analysis~\cite{Middleton:2016:GGS:2915200.2842604}
to the study of urban social behavior~\cite{Mueller2017,cranshaw2012livehoods}.
Specifically, sentiment analysis expressed by users in social networks has several applications, since a better understanding of their opinions about specific products, brands, places, or events can be useful in decision making.

\begin{figure}[h]
\centering
\subfloat["Whitesox parade"]
            {\includegraphics[width=.23\textwidth]{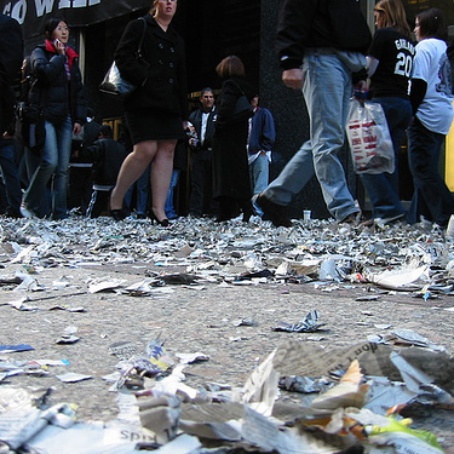}}\,
  \subfloat["Emmanuel Presbyterian"]
            {\includegraphics[width=.23\textwidth]{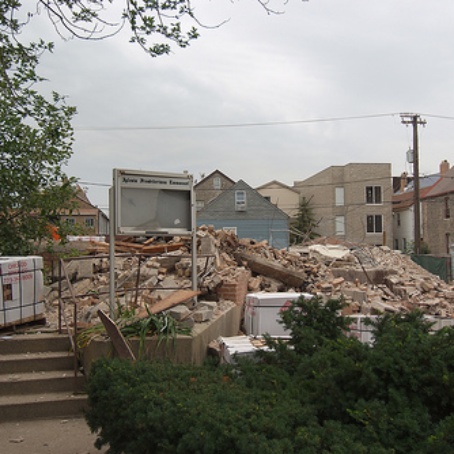}}\,
\subfloat[No label]
            {\includegraphics[width=.23\textwidth]{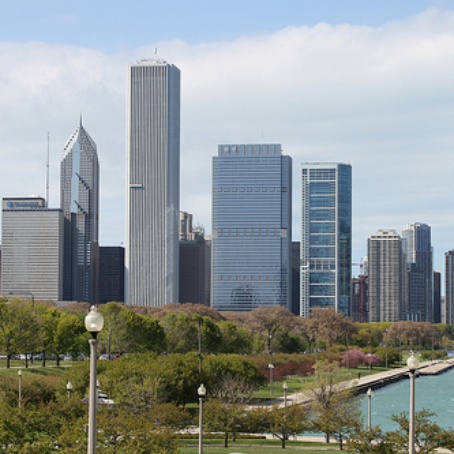}}\,
\subfloat["Smart Bar and Homeward"]
            {\includegraphics[width=.23\textwidth]{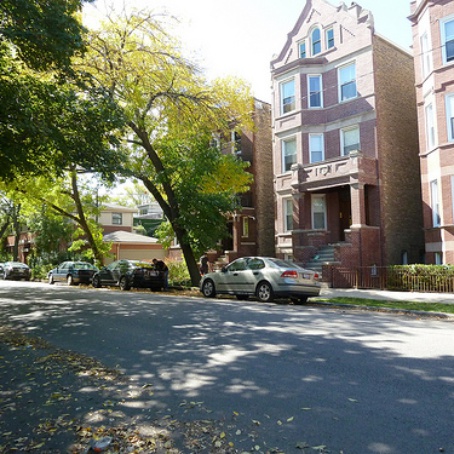}}
\caption{Examples of outdoor photos shared on Flickr and the labels (free-text) given by the users. The labels do not correctly convey the sentiment visually.}
\label{figExMotivation}
\end{figure}

Although sentiment analysis in textual content has already been developed considerably, its use in visual means is a hot-trend topic of research~\cite{campos2017pixels,cnn1,cnn4}, inspired by the fact that image sharing has become prevalent~\cite{you2015robust, imageMinute2016}. The development of novel techniques for this purpose may complement text-based approaches~\cite{Ribeiro2016, Benevenuto2015,Zhou:2017:MMD:3133943.3091995,Zhan:2019:LMD:3306215.3309543}, even those that use deep learning~\cite{Huang:2017:ESK:3026478.3052770,Baly:2016:MMH:2986034.2950050}, as well as, enable new services on platforms where the shared content is predominantly visual, such as Instagram, Snapchat, and Flickr\footnote{https://www.instagram.com; https://www.snapchat.com and https://www.flickr.com.}.
For the images illustrated in Figure~\ref{figExMotivation}, shared on Flickr, if one considers just the associated textual content, important characteristics of the sentiment visually expressed by users are not captured correctly.

The present study focus on the sentiment understanding of urban outdoor images shared by users on social networks. They can carry valuable information about urban areas since indoor images tend not to reflect specific characteristics of these scenarios (establishments in the same area probably have very different internal appearances). Therefore, it is important to explore the performance of the sentiment analysis techniques in the specific context of outdoor areas to further allow high-level tasks such as the semantic classification of urban areas, for example.

Based on the remarkable advances that deep features~\cite{LecBen15} are providing in several areas~\cite{LecBen15}, they are used in this work for sentiment classification of urban outdoor images. Five different convolutional neural network (ConvNet) architectures were compared, four widely used in machine learning, and one specifically designed for sentiment analysis~\cite{you2015robust}.  We also explored the use of semantic features, derived from SUN (Scene UNderstanding)~\cite{SUN1,SUN2} and YOLO (You Only Look Once)~\cite{Redmon_2017_CVPR} attributes, both designed for high-level scene understanding. 

In summary, our main goals are to: (i) study the impact on classification performance in each architecture by taking into account jointly deep and semantic features; (ii) investigate whether images out of context, indoor images in our case, help in the classification results; (iii) analyze what is the impact of different datasets in the results; and (iv) demonstrate an applicability of the results. 

In summary, the main contributions of this work are: 

\begin{itemize}

\item The proposal of a novel dataset of geolocalized urban outdoor images, extracted from Flickr, and labeled as positive, negative, or neutral by at least five different volunteers. Although some initiatives do not consider the neutral polarity~\cite{you2015robust,cnn1,sentibank}, we chose to use it to contemplate cases where an image is not associated with extreme polarities, as well as cases where there is no consensus between users about the sentiment expressed. We believe its introduction leads to results closer to reality since the volunteers of our research classified a significant amount of images as neutral.

\item The proposal of a framework for sentiment analysis that combines global scene features and high-level semantics. As part of this work, we evaluated four state-of-the-art widely used ConvNet architectures, namely VGG16~\cite{SimonyanZ14a}, ResNet50~\cite{He_2016_CVPR}, InceptionV3~\cite{7780677}, and DenseNet169~\cite{huang2017densely}, all fine-tuned with pre-trained ImageNet weights~\cite{imagenet_cvpr09}, and the state-of-art architecture of You~\etal~\cite{you2015robust}, designed specifically for sentiment analysis. 

\item An analysis of the impact of merging deep features with semantic features derived from SUN~\cite{SUN1,SUN2} and YOLO~\cite{Redmon_2017_CVPR} attributes, both initially designed for high-level scene understanding in terms of categories to describe scenes and objects on images, respectively. 
We carried out experiments in different datasets and concluded that the incorporation of semantic features in the models improved the accuracy result of previous initiatives for the context studied. Furthermore, we also show that
the use of semantic attributes improved the performance of all ConvNets architectures but had a much more significant impact on the most straightforward architectures.

\item A cross-dataset generalization study to evaluate the robustness capacity of the proposed architecture when considering different datasets for training and testing. We found that the simplest architecture evaluated, the ConvNet of You~\etal~\cite{you2015robust}, demonstrated to be more robust in all considered scenarios. We also observed that image edition has an influence on the robustness performance, e.g., Instagram images, unlike Flickr images, tend to have filters, written messages, and other resources. Besides, we have an indication that the dataset size plays a positive role in this aspect.

\item The evaluation of the influence of indoor images in the sentiment classification of outdoor images. By performing different sets of experiments, we observed that there is no significant gain in using indoor images that pay of the high extra cost demanded.

\item The demonstration of the feasibility of our approach in a real-world scenario. We analyzed the sentiment in outdoor images shared on Chicago through Flickr -- new images were obtained for this analysis. This evaluation shows that our approach could be useful in understanding the subjective characteristics of areas and their inhabitants. The results suggest that particular areas of the city tend to concentrate more images of a specific class of sentiment, having predominant inherent characteristics. 
\end{itemize}

The ConvNet architectures, trained models, and the proposed outdoor image dataset are publicly available at \href{http://dainf.ct.utfpr.edu.br/outdoorsent}{http://dainf.ct.utfpr.edu.br/outdoorsent}. The remainder of this paper is organized as follows. Section~\ref{sectrabrel} reviews the literature on sentiment analysis. Section~\ref{SecMetodolo} details the methodology used in this study.  Section~\ref{secReultados} discusses the results obtained. Section~\ref{secCrossDataset} shows the results when evaluating the robustness capacity of the proposed architecture when considering different datasets for training and testing (cross-dataset generalization). Section \ref{secApplication} presents an evaluation of our results in a real-world scenario. Section~\ref{secLimitacoes} presents the potential implications and limitations of the study. Finally, Section~\ref{secConclusion} concludes the study.

\clearpage
\section{Related Work}\label{sectrabrel}

Automated sentiment analysis is essential for several tasks, including those related to the understanding of human behavior and decision-making support. For instance, Kramer~\etal~\cite{kramer2014} suggested that emotions expressed on Facebook can be transferred to other people without their awareness, which could cause large-scale emotional contagion in virtual communities. Choudhury~\etal~\cite{de2013predicting} estimated the risk of depression by analyzing behavioral attributes such as social engagement, emotion, language, and linguistic styles in tweets from users diagnosed with clinical depression. Golder and Macy~\cite{golder2011} found that positive and negative emotions expressed on Twitter match well-known daytime and seasonal behavioral patterns in different cultures. 

Algorithms for this active field of research are mostly concentrated on textual content \cite{Baly:2016:MMH:2986034.2950050,Huang:2017:ESK:3026478.3052770,conf/interspeech/MikolovKBCK10,doi:10.1162/neco.1997.9.8.1735}. However, there are some limitation on exploring only this source of data for sentiment analysis for various scenarios, mainly where there is limited text, and the content is primarily visual, such as Instagram posts \cite{Ji2016}. 

Regarding sentiment analysis on visual content, e.g., images and videos, algorithms for this task, traditionally, were based on low-level visual attributes such as colors~\cite{visual1}, texture~\cite{visual2}, image gradients~\cite{visual3}, metadata and speech transcripts~\cite{visual4}, and descriptors inspired by psychology and art theory~\cite{visual5}. There are also studies dedicated to assess the impact of high-level abstraction for sentiment analysis, such as attributes on visual content regarding, for instance, material (e.g., metal), surface (e.g., rusty), as explored by Yuan et al. \cite{sentribute}, and visual concepts that are strongly related to sentiments, as used by Borth et al. \cite{sentibank}.

In recent years, inspired by the breakthroughs of convolutional neural networks (ConvNet)~\cite{LecBen15} in machine learning, many authors proposed novel architectures for visual sentiment analysis. For example, You~\etal~\cite{you2015robust} used a probabilistic ConvNet sampling to reduce the impact of noisy data by removing training instances with similar sentiment scores. Chen~\etal~\cite{cnn2} used transfer learning from ImageNet weights~\cite{imagenet_cvpr09} in their ConvNet to deal with biased training data, which only contains images with strong sentiment polarity. Discriminative face features~\cite{cnn9}, extracted from ConvNets architectures, were also used to estimate people's happiness in the wild. Cai and Xia~\cite{cnn1} proposed to explore ConvNet to combine visual features with sentiment concepts, automatically identified from the tags available on public images on the Web for detecting sentiments depicted in the images. Song et al. \cite{SONG2018218} proposed to include visual attention on images into ConvNets aiming to boost the sentiment classification performance. 

In the same direction, some approaches consider the joint representation over multi-modal inputs from microblogs containing video, image, text, and emoticons to sentiment analysis. For instance, Chen et al. \cite{chen2015} presents a multi-modal classification model for microblog sentiment classification, taking into account the correlation and independence among distinct modalities. Poria~\etal~\cite{PORIA2017217} proposed an approach to extract characteristics from textual and visual modalities using ConvNets. You et al. \cite{You2015MM} introduced a model considering ConvNet that takes into account the modality, text, and image, to classify sentiments. To tackle some of the problems of previous approaches, especially the scalability issue related to human-labeled sentiment data, some approaches propose the exploration of emotion related symbols, such as emoticons, common in specific systems, such as Twitter and Sina Weibo\footnote{A Twitter-like system from China.} \cite{LIN2018258,8052551}. Similarly, Vadicamo~\etal~\cite{8265255} presented an approach that considers content composed of text and image, e.g., tweets, to extract labels referring to sentiment polarity. Araque~\cite{ARAQUE2017236}  proposed to fuse deep and classic hand-crafted features for sentiment analysis of textual data in social applications by using an ensemble of classifiers.

In a similar direction, Ortis et al. \cite{8516481} compared text provided by users to label social images (subjective) and text automatically extracted from the visual content using four deep learning models trained to perform visual inference tasks (objective). They considered different combinations of subjective text, objective text, and visual features for sentiment polarity estimation using an SVM-based image sentiment classifier. One important conclusion is that subjective text introduces noise and affects classification performance. On the other hand, the combination of visual features and objective text produced better results. 

While there are some initiatives showing the importance of outdoor images to the semantic classification of urban areas, such as the study of Santani~\etal~\cite{Santani2018} that extracts labels for outdoor urban images, none of the previous studies focused on sentiment analysis of urban outdoor images.

In this study, in addition to a comparative study of different ConvNets already consecrated in the area of machine learning, we proposed a novel framework that incorporates into the classification process semantic features extracted directly from the images (without using metadata). Also differentiating from previous studies, we analyzed the impact of considering indoor images in the classification of outdoor ones. Then, we evaluated the cross-dataset generalization for different architectures in different scenarios. 

\section{Methodology}\label{SecMetodolo}

\subsection{Overview}

Sentiment analysis based on images has an inherent subjectivity since it involves the visual recognition of objects, scenes, actions, and events. Therefore, learning approaches need to be robust to be able to cover different domains.

\begin{figure}[!htb]
\input{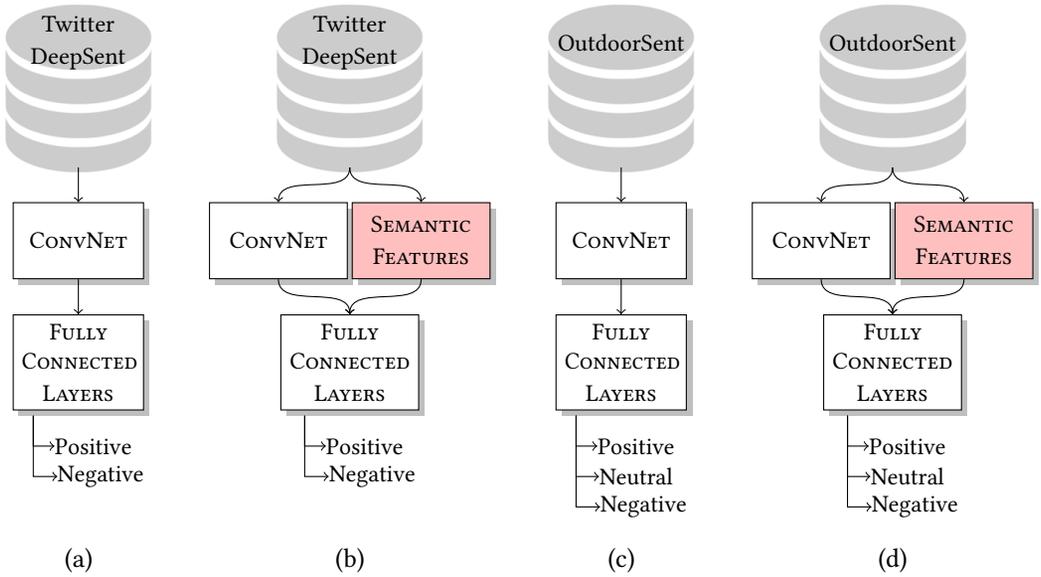}
\caption{Sentiment analysis framework. We considered four distinct experimental setups for the proposed framework: (a) sentiment analysis of indoor and outdoor images from Twitter dataset (DeepSent), which were manually labeled as positive or negative; (b) classification of Twitter dataset (DeepSent) incorporating SUN and YOLO attributes, which were designed for high-level scene understanding and object recognition, respectively; (c) sentiment analysis on outdoor images from OutdoorSent, which were manually labeled as positive, neutral, or negative; and (d) sentiment analysis on OutdoorSent with SUN and YOLO attributes.}
\label{fig:architectures}
\end{figure}

In this context, we first evaluate four different experimental setups, illustrated in Figure~\ref{fig:architectures}, with varying datasets of images, as well as with and without combining the activation maps of the convolution layers with SUN~\cite{SUN2} and YOLO~\cite{Redmon_2017_CVPR} semantic attributes. 
SUN attributes represent the principal categories used by people to describe scenes and are intended to be an intermediate representation used in applications such as scene classification, search based on semantics, and automatic labeling, to mention some examples. YOLO attributes, more specifically those extracted using the YOLO9000 network, include many objects of interest for urban sentiment analysis such as guns, fire, homeless, etc. The detection of such semantic attributes incorporates knowledge useful to perform visual inference tasks.

The first two experimental setups (Figures~\ref{fig:architectures}(a)-(b)) are based on a publicly available image dataset (DeepSent) where only positive and negative polarities are considered. The last two (Figures~\ref{fig:architectures}(c)-(d)) use the novel urban outdoor image dataset proposed in this work (OutdoorSent), a dataset that takes into account an additional polarity: neutral. Section \ref{secData} details these datasets.

Each setup compared five ConvNet architectures, whose performance has excelled in different areas: VGG16~\cite{SimonyanZ14a}, Resnet50~\cite{He_2016_CVPR}, InceptionV3~\cite{7780677} and DenseNet169~\cite{huang2017densely}, fine-tuned with pre-trained ImageNet weights~\cite{imagenet_cvpr09}, and the state-of-art architecture of You~\etal~\cite{you2015robust}, designed specifically for sentiment analysis. 

Section \ref{secArchi} presents the proposed ConvNet framework for sentiment analysis in outdoor images. Next, we evaluate whether indoor images can help in the classification of outdoor images. In addition, we investigate the cross-dataset generalization power of the proposed framework. 

\subsection{Datasets}\label{secData}

In order to evaluate the selected ConvNets and semantic attributes, we used datasets with images from different domains. Specifically, we considered one dataset composed of indoor and outdoor images (Section \ref{secDeepSent}), and another one containing only outdoor images (Section \ref{secOutdoorSent}). All images are in the JPEG format.

\subsubsection{Twitter dataset - DeepSent}\label{secDeepSent}

The first dataset used in this work, known as DeepSent, consists of 1,269 images of Twitter and is available in~\cite{you2015robust}. All samples were manually labeled as positive or negative by five people using the Amazon Mechanical Turk\footnote{https://www.mturk.com.} (AMT) crowd-sourcing platform.

The dataset is subdivided according to the consensus of the labels, that is, taking into account the number of people that attributed the same sentiment to a given image. Table~\ref{tab:base_twitter} details the distribution of the results, where ``five agree'' indicates that all five AMT workers labeled the image with the same sentiment, ``four agree'' demonstrates that at least four gave the same label and ``three agree'' that at least three agreed on the rating.

 \begin{table}[!htb]
\centering
\begin{tabular}{|c|c|c|c|}
\hline
\textbf{Sentiment} & \textbf{\textit{Five agree}} & \textbf{\textit{Four agree}} & \textbf{\textit{Three agree}} \\ \hline
Positive            & 581                 & 689                 & 769                  \\ \hline
Negative            & 301                 & 427                 & 500                  \\ \hline
\textbf{Total:}     & 882                 & 1116                & 1269                 \\ \hline
\end{tabular}
\caption{Image distribution in classes and subsets of DeepSent dataset~\cite{you2015robust}.}
\label{tab:base_twitter}
\end{table}

\begin{figure}[!htb]
 \subfloat[]{\includegraphics[width=3cm,height=0.18\textheight]{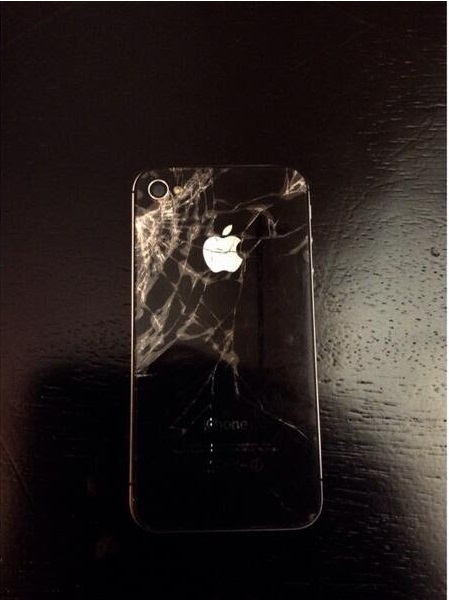}}\,
 \subfloat[]{\includegraphics[width=3cm,height=0.18\textheight]{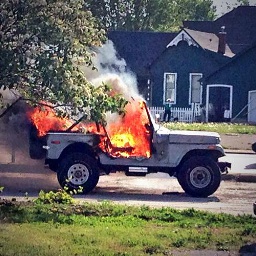}}\,
 \subfloat[]{\includegraphics[width=3cm,height=0.18\textheight]{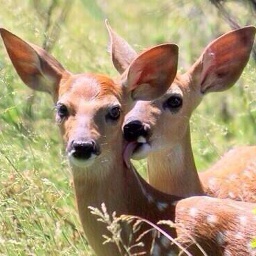}}\,
 \subfloat[]{\includegraphics[width=3cm,height=0.18\textheight]{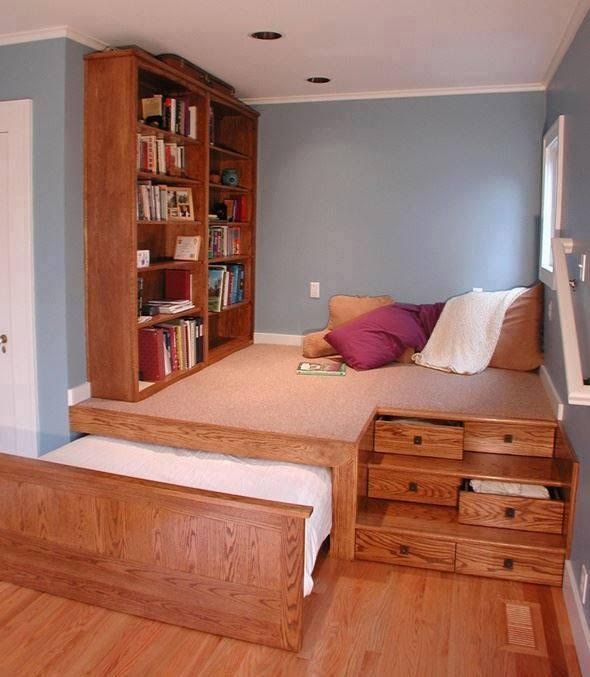}}
 \caption{Examples of negative (a-b)  and positive (c-d) images from DeepSent.}
 \label{fig:base_twitter}
\end{figure}

Figure~\ref{fig:base_twitter} shows examples of images labeled as positive and negative by volunteers of the DeepSent dataset. Note that one could argue that some of the images are more neutral than the sentiment classified.

\subsubsection{OutdoorSent}\label{secOutdoorSent}

The dataset proposed in this study, called OutdoorSent\footnote{ http://dainf.ct.utfpr.edu.br/outdoorsent}, is composed only by outdoor images. We aim to facilitate studies that demand more representative characteristics of different urban areas, considering that indoor images of establishments in the same area can be quite different.

For this dataset, we take into account 40,516 publicly available Flickr images, all of them are geolocalized in the city of Chicago. From that initial dataset, we select only the 19,411 classified as outdoor by the ConvNet Places365~\cite{zhou2017places} pre-trained for the Places2 dataset (a repository of eight million images). Precisely for this task, the Residual Network architecture (WideResNet18) was used because it has obtained the best classification accuracy and speed.

To create a training dataset, a subset of 1,950 images were randomly selected. Each of them was labeled based on the evaluation of at least five different volunteers, who graded each image according to the sentiment it represents: 1 (negative), 2 (slightly negative), 3 (neutral), 4 (slightly positive), and 5 (positive). Since we use five different classes for the characterization of sentiment, the subdivision of the dataset according to the consensus is not feasible. Thus, the final label was defined based on the grade average:

\begin{itemize}
\item Average below 2.2: negative;
\item Average between 2.2 and 3.8: neutral;
\item Average above 3.8: positive.
\end{itemize}

Each of the 30 volunteers (mostly undergraduate students) labeled 300 images divided into blocks of 15 images to make the process less tiring and error-prone (since users could respond to each block at different times). The interface was built using GoogleForms, with each block of images explaining the purpose of the project and the methodology that should be used, accompanied by an example. The choice of the images that would be shown in each block was random, and the forms were generated automatically with the aid of a script. Thus, it was possible to guarantee that each form contained images without repetitions. Each volunteer received 15 distinct forms, and the responses for each user were counted only once. Using internal management, we ensure that at least five different volunteers answered each form.

Table~\ref{tab:base_chicago} details the class distribution. The difference between the number of samples in the different classes is due to the nature of the images of the dataset, which has more neutral and positive images.

\begin{table}[!ht]
\centering
\begin{tabular}{|c|c|}
\hline
\textbf{Sentiment} & \textbf{Quantity of images} \\ \hline
Negative            & 259         \\ \hline
Neutral             & 1187            \\ \hline
Positive            & 504             \\ \hline
\textbf{Total:}     & 1950                       \\ \hline
\end{tabular}
\caption{Image distribution for the OutdoorSent dataset.}
\label{tab:base_chicago}
\end{table}

Besides, it should be emphasized that the inclusion of the neutral class aims to make the classification more realistic. In addition to allowing people to classify images where it is not possible to attribute a positive or negative sentiment, it enables a very subjective task to be relaxed, since different people may have a different opinion about the same image. 
It is essential to mention that the volunteers of our research classified a significant amount of images as neutral, demonstrating its relevance.
 
\begin{figure}[!htb]
 \subfloat{\includegraphics[height=0.165\textheight]{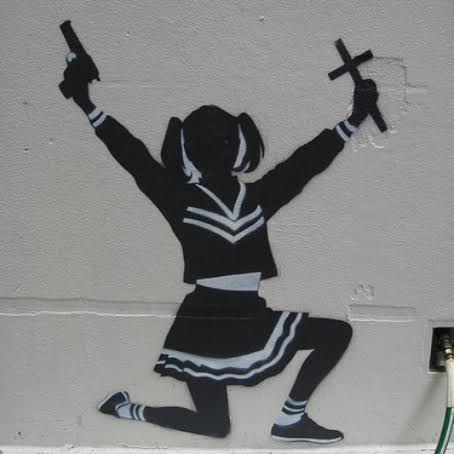}}\,
 \subfloat{\includegraphics[height=0.165\textheight]{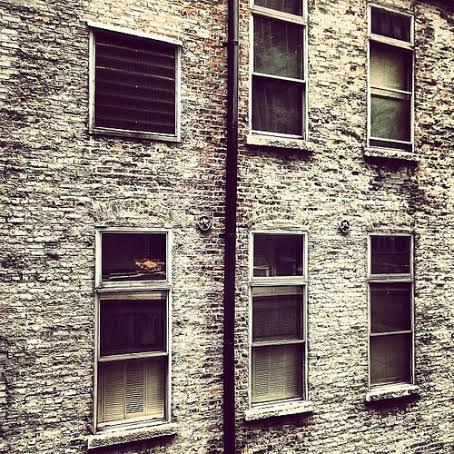}}\,
  \subfloat{\includegraphics[height=0.165\textheight]{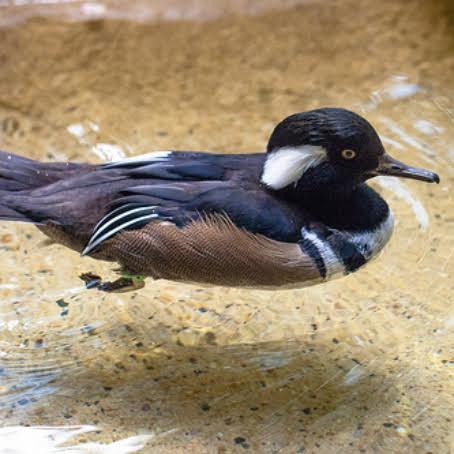}}\\ \setcounter{subfigure}{0}
  \subfloat[]{\includegraphics[height=0.165\textheight]{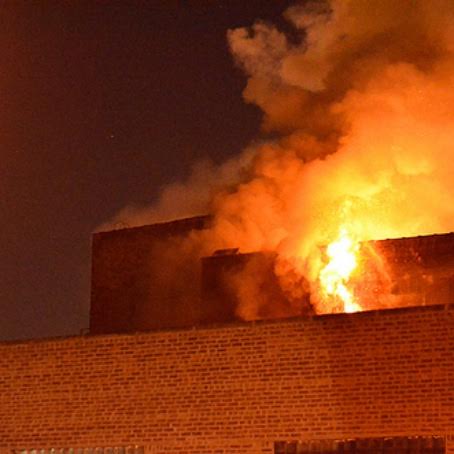}}\,
  \subfloat[]{\includegraphics[height=0.165\textheight]{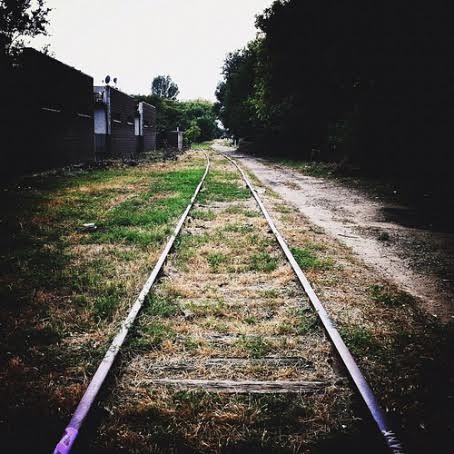}}\,
 \subfloat[]{\includegraphics[height=0.165\textheight]{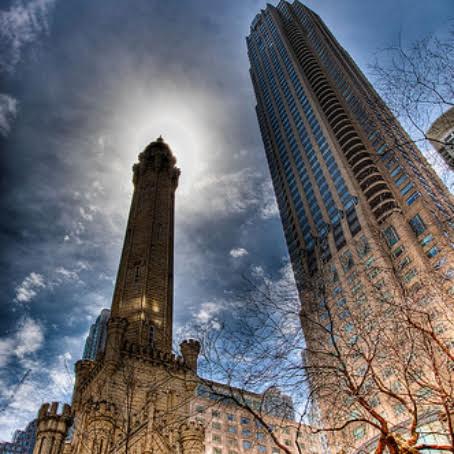}}
 \caption{Examples of images from the OutdoorSent dataset: (a) classified as negative, (b) classified as neutral, and (c) classified as positive.}
 \label{fig:base_chicago}
\end{figure}

Although previous studies have considered the neutral sentiment, in some cases, it is attributed to inherently negative scenes, such as photos of fatal accidents that show the victim~\cite{ahsan2017towards}. Due to that, our dataset is more representative.
Figure~\ref{fig:base_chicago} illustrates examples of images labeled as positive, neutral, and negative.

\subsection{ConvNet Framework}\label{secArchi}

Most state-of-the-art architectures for visual content sentiment analysis are shallow, in terms of convolutional layers, to extract deep features that could represent human sentiments. For instance, You~\emph{et al.}~\cite{you2015robust} in 2015, used a ConvNet with only two convolutions in a seminal work for the field. However, as observed by Le Cun \emph{et al.}~\cite{LecBen15}, deep convolutional neural networks are essential for the learning process of complex structures because the first convolutional layers typically represent low-level features such as edges, orientations and spatially, while deep layers combine these coarse features to recognize complex structures. 

\begin{figure*}[!htb]
  \input{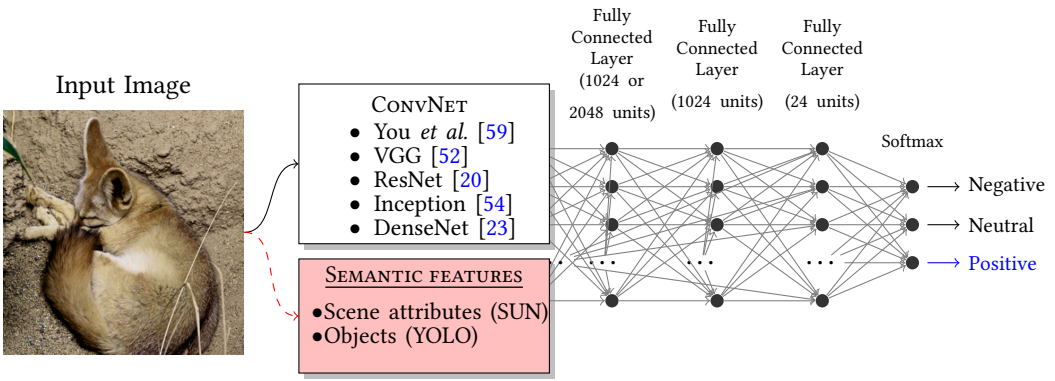}
  \caption{Proposed framework for sentiment analysis: the convolution maps and SUN + YOLO attributes are connected in a sequence of dense layers for the extraction of negative and positive sentiments (and neutral for OutdoorSent dataset). The architectures considered were: You et al.~\cite{you2015robust}, VGG, ResNet, Inception and DenseNet. The number of states in the activation layers was 1024 for You \emph{et al}.~\cite{you2015robust} due to the size of ConvNet, and 2048 for the others. } 
  \label{fig:cnn_arq}
\end{figure*}

Therefore, we consider in our framework deep state-of-the-art ConvNet architectures widely used in machine learning: VGG-16~\cite{SimonyanZ14a} (16 layers), Resnet~\cite{He_2016_CVPR} (50 layers), InceptionV3~\cite{7780677} (42 layers) and DenseNet~\cite{huang2017densely} (169 layers). These architectures were fine-tuned with pre-trained ImageNet weights~\cite{imagenet_cvpr09} to speed up the training phase. We also considered the architecture of You~\emph{et al.}~\cite{you2015robust} (2 convolutional layers) since it was explicitly designed for sentiment analysis. The ConvNet framework we developed for sentiment analysis of outdoor images is shown in Figure~\ref{fig:cnn_arq}, where the convolution maps and SUN + YOLO attributes are connected in a sequence of dense layers for the extraction of negative and positive sentiments (and neutral for OutdoorSent dataset).

\begin{figure}[!htb]
 \subfloat[]{\includegraphics[width=0.40\textwidth,height=0.18\textheight]{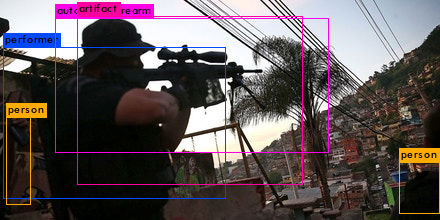}} \hspace{8pt}
 \subfloat[]{\includegraphics[width=0.40\textwidth,height=0.18\textheight]{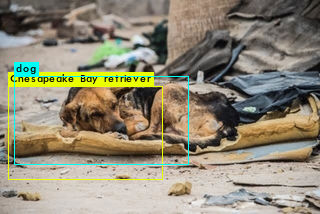}}
 \caption{Examples of semantic knowledge associated to urban outdoor scenes found by the YOLO9000 detector: (a) a firearm and ordinary persons may be a good indicator of a negative sentiment; (b) the detection of a single dog does not bring sufficient knowledge to extract the sentiment of an image, thus, the context explored by global features plays an essential role.}
 \label{fig:challenges}
\end{figure}

Another key aspect of our framework is the merging process of convolutional activation maps with high-level semantic features, which are crucial to tackling some significant challenges of sentiment analysis concerning visual content. For an example of how the existence of some objects in the scene can be decisive in describing the scene polarity see Figure~\ref{fig:challenges}~(a), which shows an image classified with YOLO attributes. However, there are cases where the detection of ordinary objects in a scene by itself might be not enough to indicate certain sentiment polarity, see Figure~\ref{fig:challenges}~(b). In this case, to minimize the problem, we must also analyze features from the global scene. Therefore, we believe that the synergistic combination of global scene features, extracted by a ConvNet architecture (deep features), and high-level semantics obtained from general-purpose detectors or specialized networks (semantic features) is a good way to tackle this problem. 

As high-level semantics, we used SUN and YOLO attributes. The first one was extracted using the ConvNet Places365~\cite{zhou2017places} pre-trained for the Places2 dataset. This feature descriptor has $102$ dimensions and is related to scene attributes such as materials, surface properties, lighting, affordances, and spatial layout. Furthermore, we used the YOLO9000 network~\cite{Redmon_2017_CVPR} for object detection because it was trained for 9,418 categories that include many objects of interest for urban sentiment analysis such as guns, fire, homeless, etc. Thus, we have in the first fully connected layer $N$ features extracted by the convolutional layers that can be joined with $102$ SUN attributes or/and $9,418$ YOLO attributes.

During the initial tests, we observed that sentiment analysis polarization into a binary classification of positive and negative images is not suitable to represent reality; therefore, we created the OutdoorSent image dataset - where the neutral label is considered. To process these images, we extended the proposed framework with a ``neutral'' state, as done by Ahsan~\emph~{et al.}~\cite{ahsan2017towards} for social event images (but not focused on outdoor images).

\section{Experiments and Results}\label{secReultados}

In this section, we compare the performance of the four experimental setups discussed in the previous section (Figure~\ref{fig:architectures}). Besides evaluating the accuracy of the results, we analyze the images classified as most likely to belong to each class. 

To achieve more significant results, we used 5-fold cross-validation, that is, each image dataset is first partitioned into k equally sized segments. Then, k iterations of training, validation, and test are performed, and within each iteration a different fold is held-out for test while the remaining k - 1 folds are used for learning (80\% of the images for training and 20\% for validation). Furthermore, to avoid the class imbalance problem during the training phase, we used a weighted optimization where the importance of each sample is proportional to the number of samples in each class. That is,  we give a higher weight to a minority class and lower weight to a majority class. 

We performed our experiments on an Intel i7-8700K 3.7GHz, 64GB RAM, with an NVIDIA Titan~Xp~GPU. For all tests, we used an Adam optimizer with an initial learning rate of 1e-4. 

\subsection{Experiments on DeepSent - Twitter Dataset}

To compare our results with those obtained by~\cite{campos2017pixels}, and~\cite{you2015robust}, we use the three subsets of the DeepSent dataset (three-agrees, four-agrees, and five-agrees). As this dataset has only two classes (positive and negative), the last layer of each neural network architecture has two neurons.

Table~\ref{tab:results_twitter} shows the accuracies obtained for each architecture considered in the experimental setups using the DeepSent dataset. The last two rows show the results for two hand-crafted algorithms based on color histograms (GCH and LCH~\cite{you2015robust}) by using the same dataset.

\begin{table}[!hbt]
\setlength{\tabcolsep}{7pt}
\centering
\begin{tabular}{|c|c|c|c|c|}
\hline
\multicolumn{2}{|c|}{Architecture}                                               & 5-agree & 4-agree & 3-agree \\ \hline
\multirow{4}{*}{\begin{tabular}[c]{@{}c@{}}You~\etal~\cite{you2015robust}\end{tabular}}
                                                        & without attributes     & $66.82 \pm  2.95$ & $62.69 \pm 1.21$ & $61.65 \pm 1.76$ \\ \cline{2-5}
                                                        & with SUN attributes    & $80.57 \pm  1.01$ & $78.03 \pm 3.08$ & $74.86 \pm 2.32$ \\ \cline{2-5}
                                                        & with YOLO attributes   & $73.86 \pm  2.84$ & $73.27 \pm 2.49$ & $68.70 \pm 3.52$ \\ \cline{2-5}
                                                        & YOLO + SUN attributes  & $\textbf{84.55} \pm  \textbf{1.99}$ & $\textbf{80.28} \pm \textbf{3.64}$ & $\textbf{76.28} \pm \textbf{3.96}$ \\ \hline
\multirow{4}{*}{VGG16}                                  & without attributes     & $61.35 \pm 14.48$ & $64.03 \pm 4.31$ & $64.90 \pm 4.88$ \\ \cline{2-5}
                                                        & with SUN attributes    & $75.22 \pm  5.68$ & $75.34 \pm 3.27$ & $65.69 \pm 4.80$ \\ \cline{2-5}
                                                        & with YOLO attributes   & $69.76 \pm  5.06$ & $69.14 \pm 4.36$ & $64.11 \pm 3.11$ \\ \cline{2-5}
                                                        & YOLO + SUN attributes  & $\textbf{79.43} \pm  \textbf{3.38}$ & $\textbf{79.29} \pm \textbf{1.27}$ & $\textbf{76.53} \pm \textbf{3.65}$ \\ \hline
\multirow{4}{*}{InceptionV3}                            & without attributes     & $86.48 \pm  2.15$ & $\textbf{83.33} \pm \textbf{3.16}$ & $78.90 \pm 2.77$ \\ \cline{2-5}
                                                        & with SUN attributes    & $85.80 \pm  1.59$ & $82.33 \pm 3.05$ & $\textbf{79.61} \pm \textbf{1.80}$ \\ \cline{2-5}
                                                        & with YOLO attributes   & $\textbf{87.73} \pm  \textbf{1.44}$ & $82.42 \pm 4.23$ & $78.98 \pm 3.08$ \\ \cline{2-5}
                                                        & YOLO + SUN attributes  & $87.50 \pm  2.01$ & $82.96 \pm 3.88$ & $78.58 \pm 1.54$ \\ \hline
\multirow{4}{*}{ResNet50}                               & without attributes     & $65.80 \pm  0.15$ & $60.35 \pm 2.06$ & $79.21 \pm 1.57$ \\ \cline{2-5}
                                                        & with SUN attributes    & $65.80 \pm  0.15$ & $64.31 \pm 4.92$ & $\textbf{81.50} \pm \textbf{1.03}$ \\ \cline{2-5}
                                                        & with YOLO attributes   & $85.00 \pm  1.33$ & $\textbf{83.76} \pm \textbf{2.36}$ & $78.89 \pm 4.00$ \\ \cline{2-5}
                                                        & YOLO + SUN attributes  & $\textbf{86.48} \pm  \textbf{3.12}$ & $83.51 \pm 2.05$ & $79.13 \pm 3.25$ \\ \hline
\multirow{4}{*}{DenseNet169}                            & without attributes     & $87.05 \pm  1.64$ & $82.06 \pm 1.33$ & $\textbf{80.95} \pm \textbf{1.43}$ \\ \cline{2-5}
                                                        & with SUN attributes    & $86.25 \pm  3.07$ & $82.42 \pm 2.88$ & $79.21 \pm 2.28$ \\ \cline{2-5}
                                                        & with YOLO attributes   & $\textbf{88.30} \pm  \textbf{1.63}$ & $\textbf{83.59} \pm \textbf{2.17}$ & $79.76 \pm 2.44$ \\ \cline{2-5}
                                                        & YOLO + SUN attributes  & $87.62 \pm  1.98$ & $81.97 \pm 2.57$ & $80.32 \pm 2.23$ \\ \hline
\multicolumn{2}{|c|}{GCH \cite{you2015robust}}                                   & 68.4    & 66.5    & 66.0    \\ \hline
\multicolumn{2}{|c|}{LCH \cite{you2015robust}}                                   & 71.0    & 67.1    & 66.4    \\ \hline
\end{tabular}
\caption{Mean accuracy and standard deviation for the DeepSent dataset.}
\label{tab:results_twitter}
\end{table}

The use of attributes improved the results of all ConvNets for the 5-agree subset. In subsets with no consensus (4-agree and 3-agree), the use of semantic attributes improved the accuracy for most of the cases. It is worth noting that the insertion of the SUN + YOLO attributes boost the accuracy of the simplest networks (You~\etal~\cite{you2015robust} and VGG16) in more than 15\% on average. However, in more complex ConvNets, the same attributes did not produce a significant impact. This is an interesting result as it shows that it is possible to use semantic attributes with simple ConvNets (e.g., You~\etal~\cite{you2015robust}) to produce a faster framework with highly competitive performance. 
Finally, note that ConvNets approaches improve the performance significantly over the hand-crafted algorithms based on the GCH and LCH descriptors.

\begin{figure}[!ht]
\centering
 \begin{tikzpicture}
\node[anchor=south west,inner sep=0] (image) at (0,0){
\includegraphics[width=0.50\textwidth]{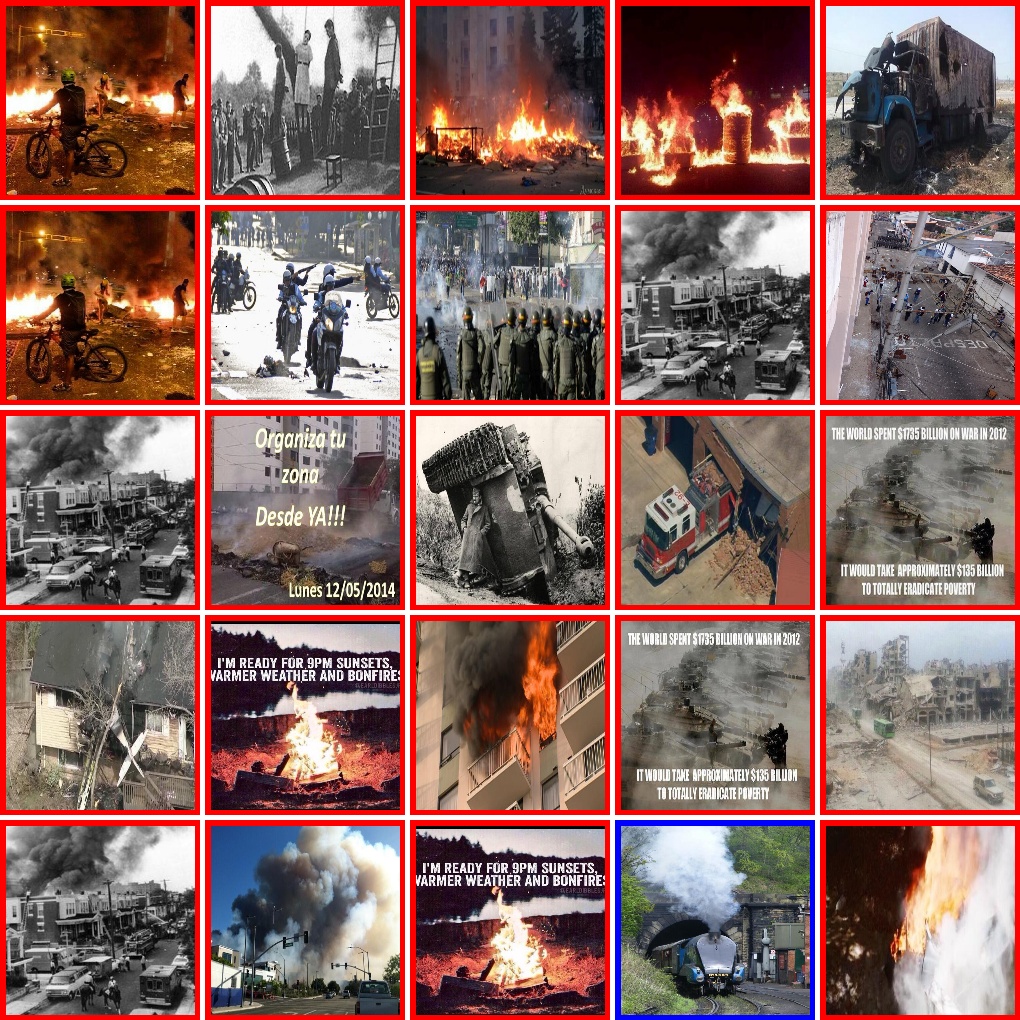}};
\node [] (A) at (2.1,-0.25) {1st (most likely to be negative)};
\node [] (B) at (5.9,-0.25) {5th};
\node at (-1.5, 6.3) (cnn01) {{(a) You et. al.}};
\node at (-1.5, 4.8) (cnn02) {{(b) VGG16}};
\node at (-1.5, 3.5) (cnn03) {{(c) InceptionV3}};
\node at (-1.5, 2.1) (cnn04) {{(d) ResNet50}};
\node at (-1.5, 0.7) (cnn05) {{(e) DenseNet169}};
\end{tikzpicture}
 \caption{Examples of images from DeepSent dataset classified as negative sentiment for different ConvNets, all using YOLO + SUN attributes: (1st row) You~\etal~\cite{you2015robust}, (2nd row) VGG16, (3rd row) InceptionV3, (4th row) ResNet50 and (5th row) DenseNet169. The images are sorted by the prediction probability given by the network in descending order. The border of the image represents its label, red to negative and blue to positive. }
 \label{fig:negTw}
\end{figure}
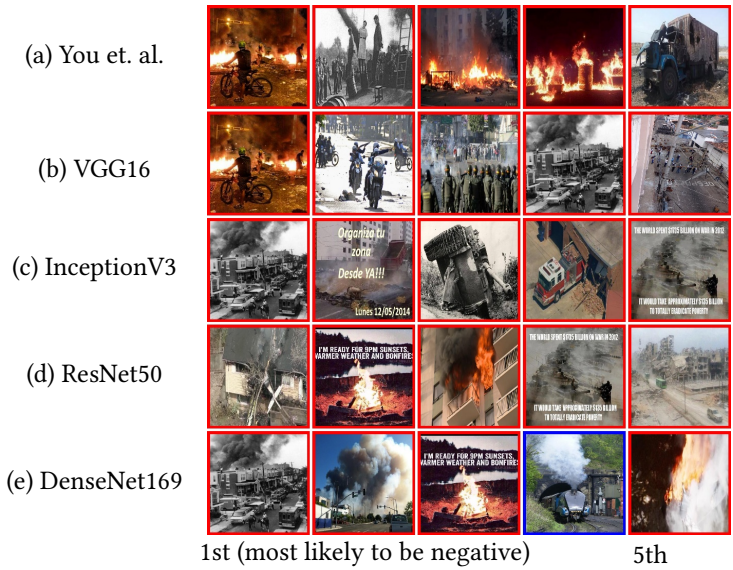

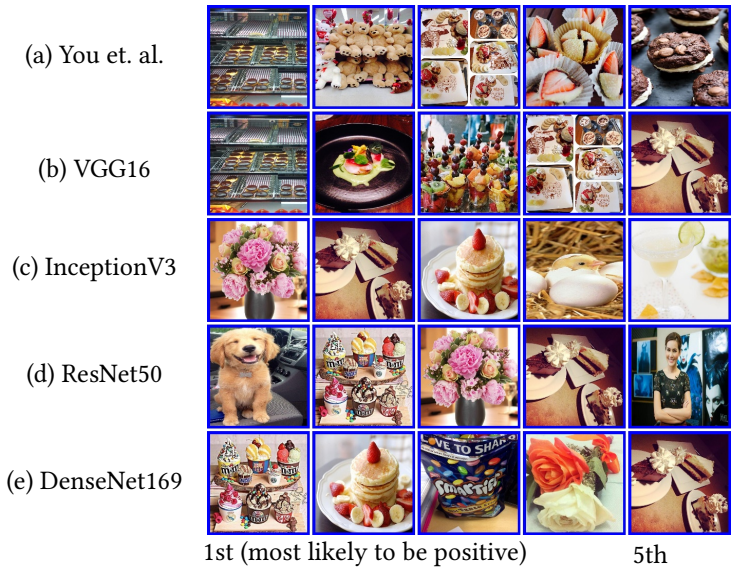
\begin{figure}[!ht]
\centering
 \begin{tikzpicture}
\node[anchor=south west,inner sep=0] (image) at (0,0){
\includegraphics[width=0.50\textwidth]{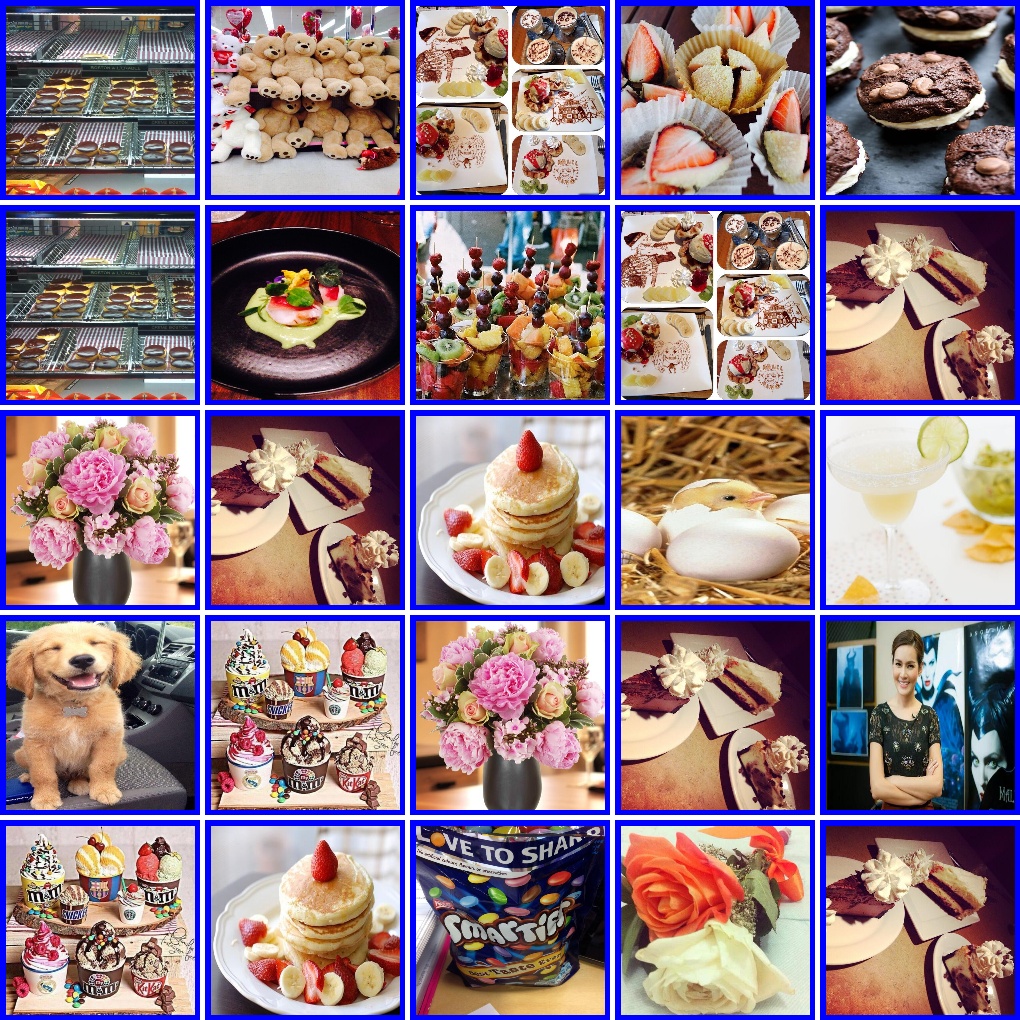}};
\node [] (A) at (2.1,-0.25) {1st (most likely to be positive)};
\node [] (B) at (5.9,-0.25) {5th};
\node at (-1.5, 6.3) (cnn01) {{(a) You et. al.}};
\node at (-1.5, 4.8) (cnn02) {{(b) VGG16}};
\node at (-1.5, 3.5) (cnn03) {{(c) InceptionV3}};
\node at (-1.5, 2.1) (cnn04) {{(d) ResNet50}};
\node at (-1.5, 0.7) (cnn05) {{(e) DenseNet169}};
\end{tikzpicture}
 \caption{Examples of images from the DeepSent dataset classified as positive sentiment for different ConvNets, all using YOLO + SUN attributes: (1st row) You~\etal~\cite{you2015robust}, (2nd row) VGG16, (3rd row) InceptionV3, (4th row) ResNet50 and (5th row) DenseNet169. The images are sorted by the prediction probability given by the network in descending order. The border of the image represents its label, red to negative and blue to positive. }
 \label{fig:posTw}
\end{figure}

For each model that makes use of the YOLO + SUN attributes, the five images that were most likely to belong to each class were selected. Figures~\ref{fig:negTw} and~\ref{fig:posTw} show the selected images to the negative and positive classes, respectively. 

As we can see on those figures, there is only one error for DenseNet169 (fifth row) that classified a positive image as negative, probably because of the train smoke. It is interesting to note that not all top-five images appear in more than one architecture for all the scenarios evaluated. This is not necessarily a problem, but just a suggestion that the learning process can be different for each considered architecture.

\subsection{Experiments on OutdoorSent}

Table~\ref{tab:results_chicago} shows the accuracy results for all ConvNets in all images of OutdoorSent dataset. As we can see, they are lower than those observed for the DeepSent dataset (the one based on Twitter). This was expected, since this dataset has three classes instead of two and, despite having only outdoor images, has a wider range of scenarios. Again, the use of semantic attributes improved the accuracy for all ConvNets but had a much more significant impact on the simplest architectures. 

\begin{table}[!ht]
     \setlength{\tabcolsep}{3pt}
    \centering
    \begin{tabular}{|c|c|c|c|c|}
    \hline
    Architecture            & without attributes & SUN attributes & YOLO attributes & YOLO + SUN attributes\\ \hline
    You~\etal~\cite{you2015robust}  & $37.87 \pm 19.39$ & $48    \pm 2.18$ & $49.38 \pm 10.41$ & \textbf{56.24} $\pm$ \textbf{6.32} \\ \hline
    VGG16                           & $32.33 \pm 23.28$ & $57.64 \pm 5.22$ & $39.06 \pm  8.57$ & \textbf{60.67} $\pm$ \textbf{1.92} \\ \hline
    InceptionV3                     & $60.71 \pm 2.77$  & $60.35 \pm 3.96$ & \textbf{61.38} $\pm$  \textbf{2.76} & $60.61 \pm 3.01$ \\ \hline
    ResNet50                        & $60.3  \pm 1.81$  & $60.4  \pm 3.51$ & $60.56 \pm  3.02$ & \textbf{61.38} $\pm$ \textbf{2.67} \\ \hline
    DenseNet169                     & $62.1  \pm 3.12$  & \textbf{63.43} $\pm$ \textbf{3.15} & $59.89 \pm  3.25$ & $62.71 \pm 3.56$ \\ \hline
    \end{tabular}
    \caption{Mean accuracy and standard deviation for the OutdoorSent dataset.}
    \label{tab:results_chicago}
\end{table}

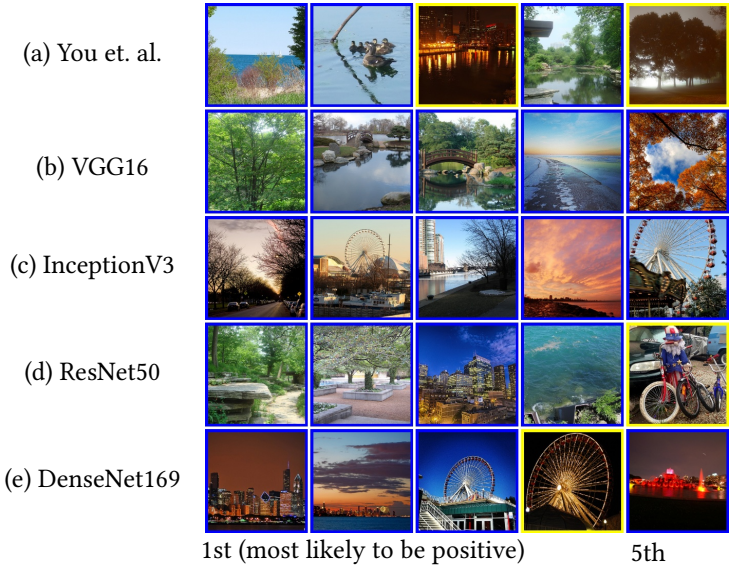
\begin{figure}[!htb]
\centering
 \begin{tikzpicture}
\node[anchor=south west,inner sep=0] (image) at (0,0){
\includegraphics[width=0.50\textwidth]{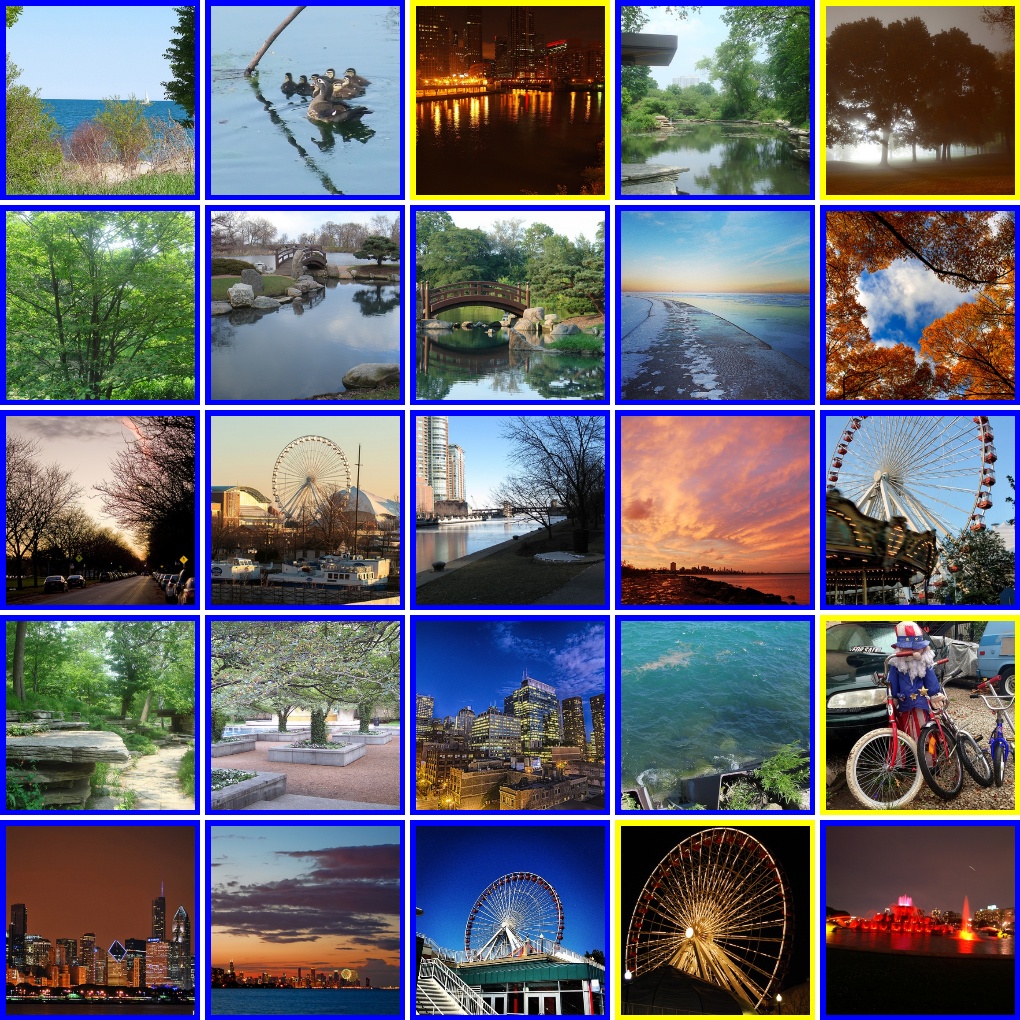}};
\node [] (A) at (2.1,-0.25) {1st (most likely to be positive)};
\node [] (B) at (5.9,-0.25) {5th};
\node at (-1.5, 6.3) (cnn01) {{(a) You et. al.}};
\node at (-1.5, 4.8) (cnn02) {{(b) VGG16}};
\node at (-1.5, 3.5) (cnn03) {{(c) InceptionV3}};
\node at (-1.5, 2.1) (cnn04) {{(d) ResNet50}};
\node at (-1.5, 0.7) (cnn05) {{(e) DenseNet169}};
\end{tikzpicture}
 \caption{Images from the OutdoorSent dataset classified as positive sentiment for different ConvNets, all using YOLO + SUN attributes: (1st row) You~\etal~\cite{you2015robust}, (2nd row) VGG16, (3rd row) InceptionV3, (4th row) ResNet50 and (5th) DenseNet169. The images are in descending order of the prediction probability given by the network. The border of the image represents its label, red to negative, blue to positive, and yellow to neutral.}
 
 \label{fig:pos}
\end{figure}

\begin{figure}[!htb]
\centering
 \begin{tikzpicture}
\node[anchor=south west,inner sep=0] (image) at (0,0){
\includegraphics[width=0.5\textwidth]{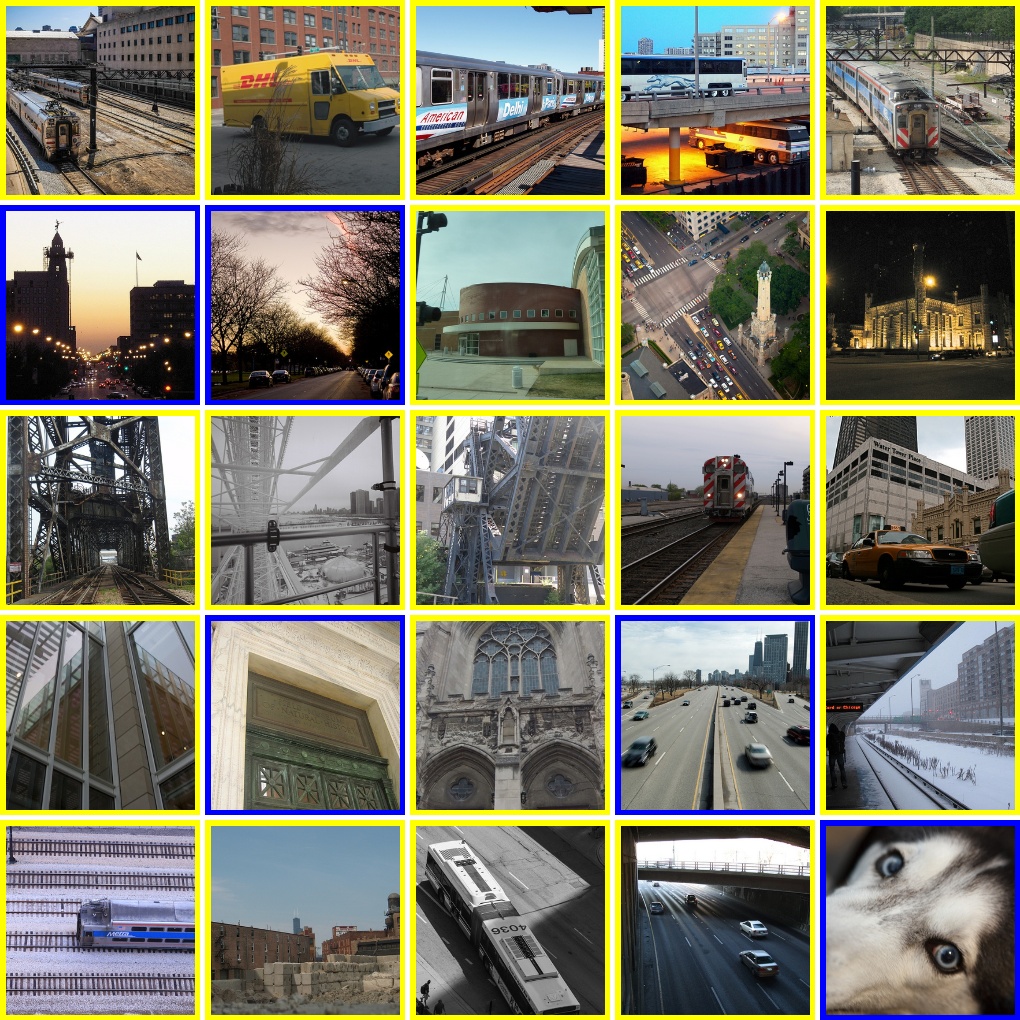}};
\node [] (A) at (2.1,-0.25) {1st (most likely to be neutral)};
\node [] (B) at (5.9,-0.25) {5th};
\node at (-1.5, 6.3) (cnn01) {{(a) You et. al.}};
\node at (-1.5, 4.8) (cnn02) {{(b) VGG16}};
\node at (-1.5, 3.5) (cnn03) {{(c) InceptionV3}};
\node at (-1.5, 2.1) (cnn04) {{(d) ResNet50}};
\node at (-1.5, 0.7) (cnn05) {{(e) DenseNet169}};
\end{tikzpicture}
\caption{Images from the OutdoorSent dataset classified as neutral for different ConvNets, all using YOLO + SUN attributes: (1st row) You~\etal~\cite{you2015robust}, (2nd row) VGG16, (3rd row) InceptionV3, (4th row) ResNet50 and (5th) DenseNet169. The images are in descending order of the prediction probability given by the network. The border of the image represents its label, red to negative, blue to positive, and yellow to neutral.}
  \label{fig:neu}
\end{figure}

\begin{figure}[!htb]
\centering
 \begin{tikzpicture}
\node[anchor=south west,inner sep=0] (image) at (0,0){
\includegraphics[width=0.5\textwidth]{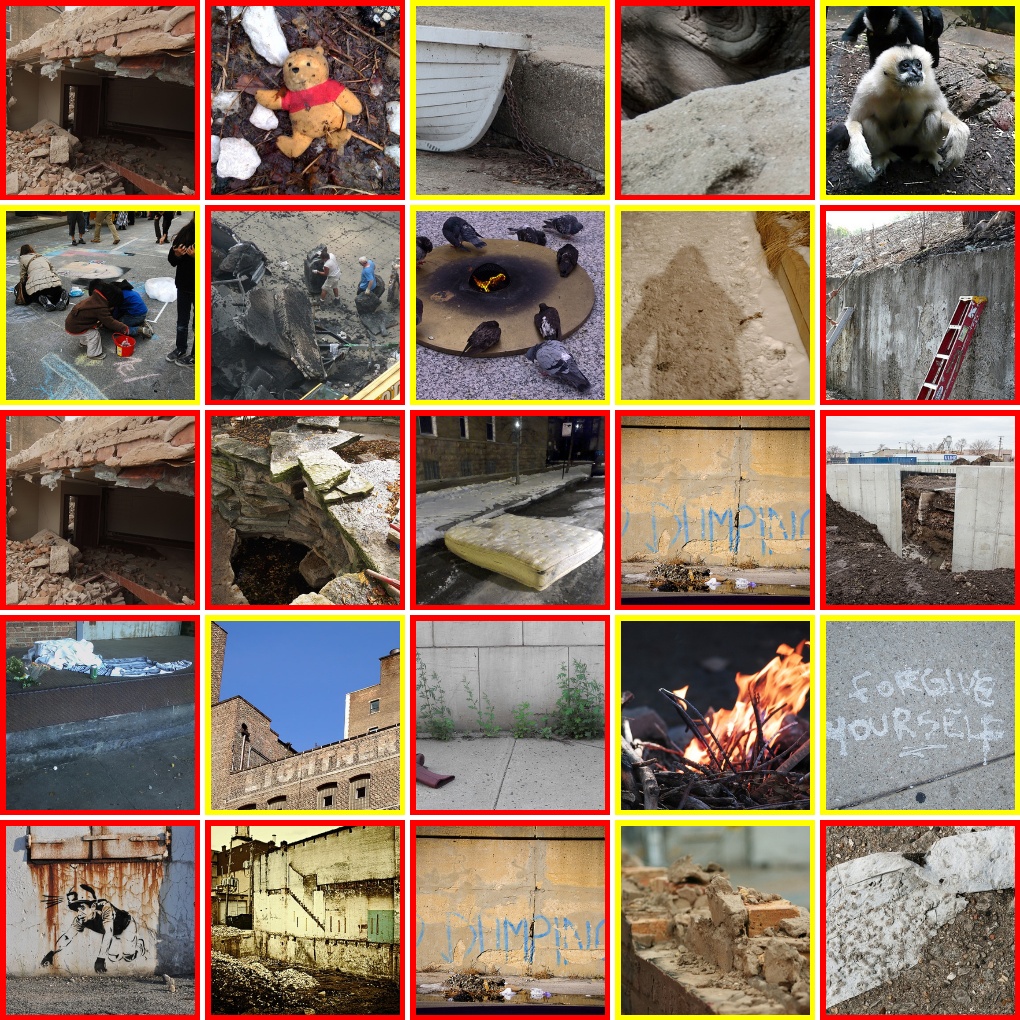}};
\node [] (A) at (2.1,-0.25) {1st (most likely to be negative)};
\node [] (B) at (5.9,-0.25) {5th};
\node at (-1.5, 6.3) (cnn01) {{(a) You et. al.}};
\node at (-1.5, 4.8) (cnn02) {{(b) VGG16}};
\node at (-1.5, 3.5) (cnn03) {{(c) InceptionV3}};
\node at (-1.5, 2.1) (cnn04) {{(d) ResNet50}};
\node at (-1.5, 0.7) (cnn05) {{(e) DenseNet169}};
\end{tikzpicture}
\caption{Images from the OutdoorSent dataset classified as negative for different ConvNets, all using YOLO + SUN attributes: (1st row) You~\etal~\cite{you2015robust}, (2nd row) VGG16, (3rd row) InceptionV3, (4th row) ResNet50 and (5th) DenseNet169. The images are in descending order of the prediction probability given by the network. The border of the image represents its label, red to negative, blue to positive, and yellow to neutral.}
\label{fig:neg}
\end{figure}

For each model that makes use of the YOLO + SUN attributes, the five images that had the highest probability of belonging to each class were selected. Figures \ref{fig:pos}, \ref{fig:neu}, and \ref{fig:neg} show the selected images to the positive, neutral, and negative classes, respectively. The images are in descending order of the prediction probability given by the network. Among these images, observe that only two are repeated, indicating that each architecture is extracting complementary features for classification. 


Considering the results for the positive class, the scenario presenting the best performance for all models, four neutral images were incorrectly included, two by You et al. \cite{you2015robust}, and one by ResNet50 and DenseNet169 each. For the case of neutral class, five positive images were wrongly included, two by VGG16, two by ResNet50, and one by DenseNet169. Turning our attention to the images of the negative class, nine neutral images were inserted (the only ConvNet that had no error in the first five images was InceptionV3).  All architectures misclassified some images, and, in general, the most problematic case was for negative instances. Note that no extreme cases were observed, i.e., a negative image being classified as positive and vice versa.

\subsection{Evaluation of Context Influence}

In this section, we envision to answer the following research question: do indoor images can help the classification of outdoor images? By investigating this question, we can gain an understanding of the possible influence of indoor images to classify outdoor images, i.e., images out of the primary scope of interest. 

To answer the stated question, we performed two rounds of experiments for each ConvNet. In the first round, we used only outdoor images in a 5-fold cross-validation experiment, where 4-folds were used for training and validation, 80\% and 20\% of the total images, respectively, and 1-fold for testing. In the second round of experiments, we repeated the first experiment, including all indoor images available for a particular dataset in the training set. Then, we evaluated the impact of this decision by using only outdoor images in the testing set (the same images from the first experiment).  

Table~\ref{tab:indoor_outdoor} present the results. As we can see, the use of indoor images to classify outdoor images contributes positively to certain networks; however, even in these cases, the gain is marginal, especially if we consider the standard deviation interval. It is worth mentioning that in these cases, we used a much larger training set for each experiment, and, thus, we increased considerably the training time (at least double) by using indoor and outdoor images.

\begin{table}[!ht]
    \centering
    \begin{tabular}{cc|c|c|}
    \cline{3-4}          &              &                        \multicolumn{2}{c|}{Testing set}   \\ 
              &              &                        \multicolumn{2}{c|}{outdoor images}   \\ \hline
  \multicolumn{1}{|c|}{Architecture}          & Training set                                        & F-score     & Accuracy   \\ \hline
    \multicolumn{1}{|c|}{\multirow{2}{*}{\begin{tabular}[c]{@{}c@{}}
                        You~\etal~\cite{you2015robust}\end{tabular}}} & outdoor images  & $45.67 \pm 2.49$ & $48.00 \pm 2.18$ \\ \cline{2-4} 
    \multicolumn{1}{|c|}{}                                            & indoor + outdoor images     & $46.60 \pm 4.48$ & $52.57 \pm 5.90$ \\ \hline
    \multicolumn{1}{|c|}{\multirow{2}{*}{\begin{tabular}[c]{@{}c@{}}
                        VGG16\end{tabular}}}                          & outdoor images  & $38.40 \pm 7.64$ & $57.64 \pm 5.22$ \\ \cline{2-4} 
    \multicolumn{1}{|c|}{}                                            & indoor + outdoor     & $25.87 \pm 9.83$ & $52.00 \pm 19.46$ \\ \hline
    \multicolumn{1}{|c|}{\multirow{2}{*}{\begin{tabular}[c]{@{}c@{}}
                        InceptionV3\end{tabular}}}                    & outdoor images  & $48.87 \pm 4.75$ & $60.35 \pm 3.96$ \\ \cline{2-4} 
    \multicolumn{1}{|c|}{}                                            & indoor + outdoor     & ${49.80} \pm {3.24}$ & ${63.58} \pm {2.95}$ \\ \hline
    \multicolumn{1}{|c|}{\multirow{2}{*}{\begin{tabular}[c]{@{}c@{}}
                        ResNet50\end{tabular}}}                       & outdoor images  & $44.47 \pm 3.1$ & $60.40 \pm 3.51$ \\ \cline{2-4} 
    \multicolumn{1}{|c|}{}                                            & indoor + outdoor     & ${45.53} \pm {5.13}$ & ${62.04} \pm {3.23}$ \\ \hline
    \multicolumn{1}{|c|}{\multirow{2}{*}{\begin{tabular}[c]{@{}c@{}}
                        DenseNet169\end{tabular}}}                    & only outdoor images  & ${48.53} \pm {2.92}$ & ${63.43} \pm {3.15}$ \\ \cline{2-4} 
    \multicolumn{1}{|c|}{}                                            & indoor + outdoor     & $47.73 \pm 4.57$ & $61.28 \pm 1.90$ \\ \hline
    \end{tabular}
    \caption{Results for the influence of indoor images on the classification of outdoor scenes. We used five cross-validations to compute the average accuracy, F-score, and standard deviation.}
    \label{tab:indoor_outdoor}
\end{table}

The next section approaches another important aspect: the robustness of the proposed architecture when considering different datasets for training and testing. 

\section{Cross-dataset generalization}
\label{secCrossDataset}
The design of architectures that are robust, i.e., able to have acceptable performance on a dataset never seen, is a trending topic on machine learning~\cite{Torralba2011,Hoffman_2018_CVPR_Workshops}, a problem also known as \emph{cross-dataset generalization}. That is, while a particular architecture can achieve a lot of progress on an individual dataset, it is common to perform worse on a different dataset of the same problem domain, and, thus, requiring substantial fine-tuning. Consequently, this process usually demands a significant amount of time and computational resources. 

\begin{figure}[!htb]
    \subfloat{\includegraphics[height=0.125\textheight]{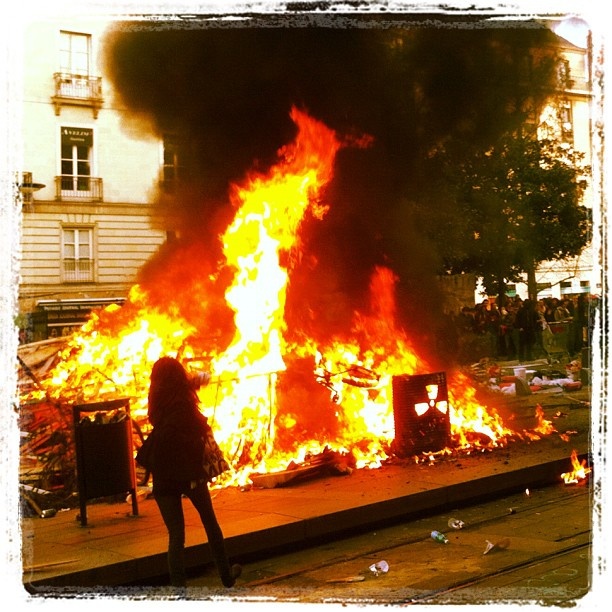}}\,
    \subfloat{\includegraphics[height=0.125\textheight]{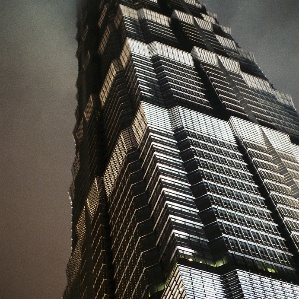}}\,
    \subfloat{\includegraphics[height=0.125\textheight]{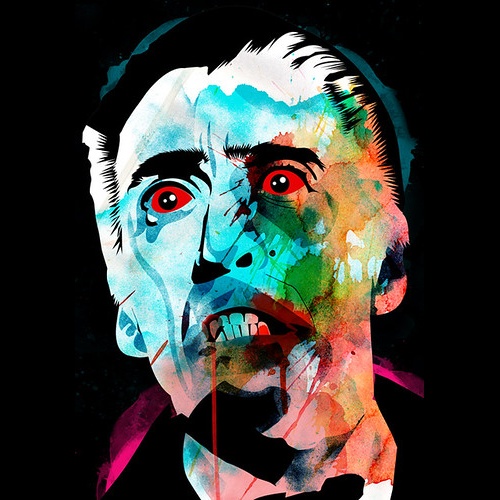}}\\

    \subfloat{\includegraphics[height=0.125\textheight]{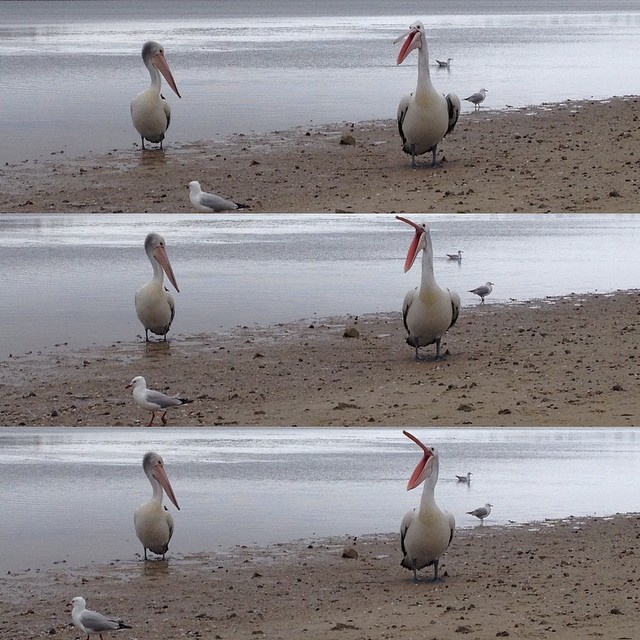}}\,
    \subfloat{\includegraphics[height=0.125\textheight]{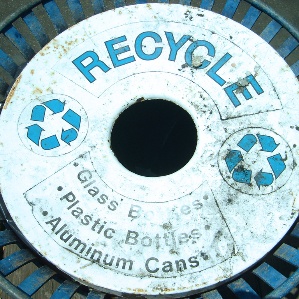}}\,
    \subfloat{\includegraphics[height=0.125\textheight]{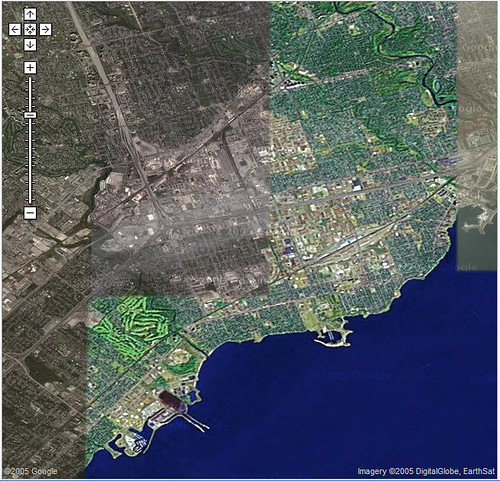}}\\\setcounter{subfigure}{0}

    \subfloat[]{\includegraphics[height=0.125\textheight]{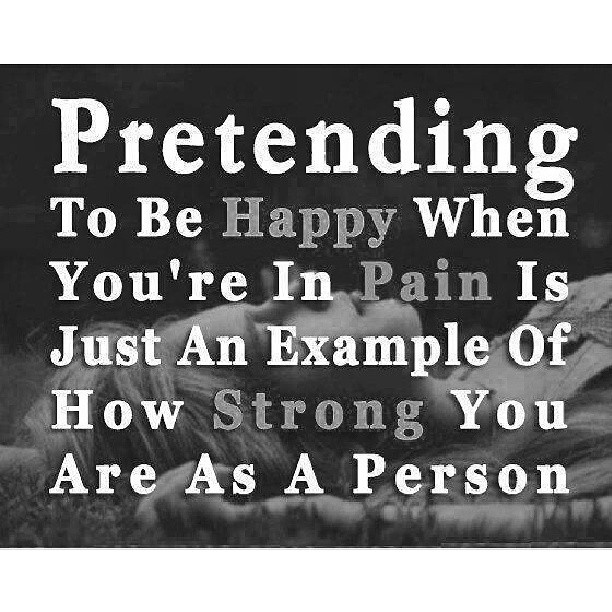}}\,
    \subfloat[]{\includegraphics[height=0.125\textheight]{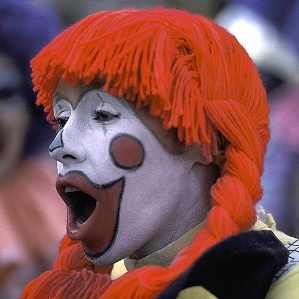}}\,
    \subfloat[]{\includegraphics[height=0.125\textheight]{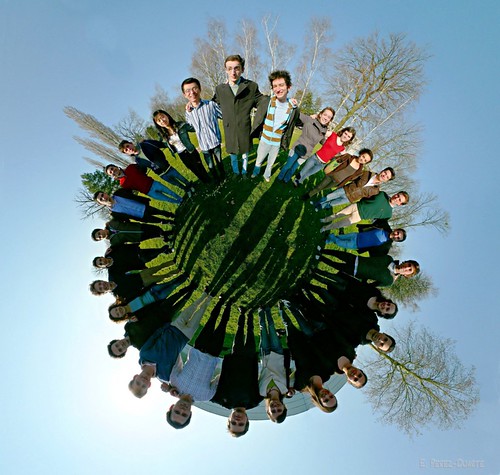}}
 \caption{Examples of images from Katsurai and Satoh~\cite{7472195} and ImageSentiment~\cite{datasetCrowdFlower} datasets, from top to bottom: Negative, Neutral and Positive. (a) Instagram-Kat, (b) Flickr-Kat, (c) ImageSentiment.}
 \label{fig:japonesa}
\end{figure}

Towards this direction, we now evaluate the robustness power of the proposed architecture when the images samples used for training and testing are from different benchmarks. For these experiments, besides of DeepSent and OutdoorSent, as detailed in Section~\ref{secData}, we considered three new benchmarks proposed specifically for image sentiment analysis:

\begin{itemize}
    \item ImageSentiment~\cite{datasetCrowdFlower}: images classified using CrowdFlower, a crowdsourcing platform. The dataset does not specify the source of the images but comprehends diverse images. In total, it has 12,884 available images (8,507 positives, 2,111 neutral, and 2,266 negatives). We separated outdoor images exploring Places, and obtained 8,001 images (5,710 positives, 1,108 neutral, and 1,183 negatives);
    
    \item Flickr-Kat: proposed by Katsurai and Satoh~\cite{7472195}. The authors classified images from Flickr using a crowdsourcing process. In total, it has 90,139 images (48,136 positives, 19,951 neutral, 12,603 negatives, and 9,449 without consensus). We separated outdoor images also exploring Places, and obtained 56,488 images (36,102 positives, 12,757 neutral, and 7,629 negatives);

    \item Instagram-Kat: also proposed by Katsurai and Satoh~\cite{7472195}. The classification process is similar, but the source is now Instagram. In total, it has 65,439 images  (33,076 positives, 17,313 neutral, 9,780 negatives, and 5,270 without consensus). We separated outdoor images exploring Places and obtained 14,801 images (8,589 positives, 3,661 neutral, and 2,551 negatives).  
   
\end{itemize}
Figure \ref{fig:japonesa} shows some examples of Instagram-kat, Flickr-Kat, and ImageSentiment. 

We first evaluated the robustness performance of our architecture by training it with all images from OutdoorSent and testing it on all other benchmarks considered. We repeat this process for all ConvNets under study. Table~\ref{tab:generalization_outdoor2} show the results considering the combination of YOLO + SUN semantic attributes. 

\begin{table}[!htb]
    \centering
  
    \begin{tabular}{cc|c|c|c|c|}
    \hline                                                     \multicolumn{2}{|c|}{\diagbox{Train}{Test}}  & ImageSentiment & Flickr-Kat & Instagram-Kat & DeepSent  \\ \hline
    \multicolumn{1}{|c|}{\multirow{2}{*}{\begin{tabular}[c]{@{}c@{}}
                        You~\etal~\cite{you2015robust}\end{tabular}}} & F-score                & \textbf{45}               & \textbf{45.67}         & \textbf{45.67}         & \textbf{63.5}      \\ \cline{2-6} 
    \multicolumn{1}{|c|}{}                                            & Accuracy               & \textbf{62.55}            & \textbf{57.49}         & \textbf{51.31}          & \textbf{57.72}      \\ \hline

    \multicolumn{1}{|c|}{\multirow{2}{*}{VGG16}}                      & F-score                & 44               & 43.67         & 44.67          & 59      \\ \cline{2-6} 
    \multicolumn{1}{|c|}{}                                            & Accuracy               & 57.72            & 54.96         & 48.73          & 47.76      \\ \hline

    \multicolumn{1}{|c|}{\multirow{2}{*}{InceptionV3}}                & F-score                & 36.67            & 39            & 37             & 45.5      \\ \cline{2-6} 
    \multicolumn{1}{|c|}{}                                            & Accuracy               & 46.84            & 46.84         & 42.19          & 32.61     \\ \hline

    \multicolumn{1}{|c|}{\multirow{2}{*}{ResNet50}}                   & F-score                & 30.67            & 33.33         & 29.67          & 30.5      \\ \cline{2-6} 
    \multicolumn{1}{|c|}{}                                            & Accuracy               & 43.59            & 45            & 38.25          & 22.08     \\ \hline

    \multicolumn{1}{|c|}{\multirow{2}{*}{DenseNet169}}                & F-score                & 28               & 30.67         & 25             & 23     \\ \cline{2-6} 
    \multicolumn{1}{|c|}{}                                            & Accuracy               & 35.95            & 39.08         & 32.05          & 14     \\ \hline

    \end{tabular}
    \caption{Robustness performance. Training using the OutdoorSent dataset with all ConNets with YOLO + SUN semantic attributes. Test against all other datasets (only outdoor images).}
    \label{tab:generalization_outdoor2}
\end{table}

As can be seen, the ConvNet of You~\etal~\cite{you2015robust} outperformed the other architectures in all scenarios evaluated. It is worth noting that this network has only two convolutional
layers. Therefore, although
well-know and much deeper architectures such as Inception, ResNet and DenseNet outperformed the ConvNet of You~\etal~when trained, evaluated and tested in the same dataset, see Tables~\ref{tab:results_twitter} and~\ref{tab:results_chicago}, they were not robust enough to generalize to unknown datasets as this simpler ConvNet did. 

Next, we performed a new set of experiments using an all-against-all dataset evaluation. In this way, we gain an understanding of the impact of a certain dataset in the robustness. In these experiments, we only evaluate the ConvNet of You~\etal~\cite{you2015robust} with YOLO + SUN semantic attributes, the one that was more robust in the previous experiment. Table~\ref{tab:generalization2} presents the results.
\begin{table}[!htb]
     \setlength{\tabcolsep}{2pt}
    \begin{tabular}{|c|c|c|c|c|c||c|}
    \hline
    \multicolumn{2}{|c|}{\diagbox{Train}{Test}}          & OutdoorSent & ImageSentiment & Flickr-Kat    & Instagram-Kat  & DeepSent   \\ \hline
    \multirow{2}{*}{OutdoorSent}                & F-score     & \textbf{57.67}$^\dagger$      & 45             & 45.67         & 45.67    & 63.5    \\ \cline{2-7} 
                                                & Accuracy    & \textbf{61.59}$^\dagger$       & 62.55          & 57.49         & 51.31 & 57.72      \\ \hline
    \multirow{2}{*}{ImageSentiment}             & F-score     & 46.33       & \textbf{59}$^\dagger$            & 48.67         & 50   & 68    \\ \cline{2-7} 
                                                & Accuracy    & 50.67       & \textbf{70.04}$^\dagger$         & 57.74         & 55.47  & 59.74    \\ \hline
    \multirow{2}{*}{Flickr-Kat}                 & F-score     & 49.67      & 50.67         & \textbf{53.33}$^\dagger$        & 49    & 64       \\ \cline{2-7} 
                                                & Accuracy    & 53.59      & 68.49         & \textbf{63.4}$^\dagger$         & 58.17     & 57.86  \\ \hline
    \multirow{2}{*}{Instagram-Kat}              & F-score     & 41          & 46             & 45.67         & \textbf{62.33}$^\dagger$    & 67.5     \\ \cline{2-7} 
                                                & Accuracy    & 40.92       & 61.58          & 57.35         & \textbf{67.73}$^\dagger$   & 65.07     \\ \hline \hline
   \multirow{2}{*}{DeepSent} & F-score   & 66$^*$                     & 65$^*$                     & 62$^*$                     & 61$^*$                     &\textbf{85}$^\dagger$ \\ \cline{2-7}
 & Accuracy  & 66.71$^*$                 & 77.93$^*$                  & 74.91$^*$                  & 71.39$^*$            &\textbf{85}$^\dagger$ \\ \hline                                            
    \end{tabular}
    \caption{Impact of different datasets in the performance. Training and testing the ConvNet of You~\etal~\cite{you2015robust} with YOLO + SUN semantic attributes for all datasets (all-against-all protocol). Only outdoor images in all tasks. Training and testing with images from the same dataset are marked with $\dagger$. The DeepSent dataset does not have neutral images; therefore, we need to exclude from all other datasets the neutral class (entries marked with $*$), as this problem is less complex (two classes), the performances were higher.}
    \label{tab:generalization2}
\end{table}

Not coming with a surprise, the best performances were achieved by training and testing with images from the same dataset. It is also worth noting that the proposed framework with the Flickr-Kat as training set presented the best cross-dataset generalization, with an average error increasing of $\approx 9.9\%$ for F-score and $\approx 6.5\%$ for accuracy. 

The worst performance was observed for the Instagram-Kat dataset and could be explained by the particular characteristics of Instagram images, making them quite different from other images under study. For example, they tend to have filters, written messages, photo sequences, and synthetic images. Filter, for instance, it is known to have an impact on the classification \cite{Chen:2015:FIC:2733373.2806348}. The other images tend to have more photos with little or no editing (some examples can be seen in Figure \ref{fig:japonesa}). 

Another fact that might contribute to the good performance using Flickr-Kat is its dataset size, which is several times bigger than the other datasets, for instance, 28 times bigger in the number of images than OutdoorSent.  
As a particular case, we included in the cross-dataset generalization experiments the DeepSent dataset, which only has images from positive and negative polarities. To make this experiment possible we made the following adaptations: (i) in the experiment ''DeepSent against other datasets'' we excluded neutral images from the other datasets since this class does not exist for the training phase; (ii) in the experiment ''Other datasets against DeepSent'' we keep the neutral class in our framework, but each image classified as neutral was considered an error. 

As expected, the performances were much higher. The reason for that is because this problem is less complex, two classes instead of three. Another reason is that images from the neutral class are hard to classify since they are often classified as positive or negative by the volunteers (they have low consensus). In this direction, it is worth noting that most misclassifications of our experiments occur associating positive and negative classes to neutral ones.

In general, the results of this cross-dataset generalization experiment are promising, given the diversity of such datasets and the simplicity of our framework (we did not use any fine-tuning for any specific dataset).


\section{Outdoor Sentiment in the City}\label{secApplication}

\begin{figure}[htb]
\centering
\subfloat[Positive]
            {\includegraphics[width=.45\textwidth]{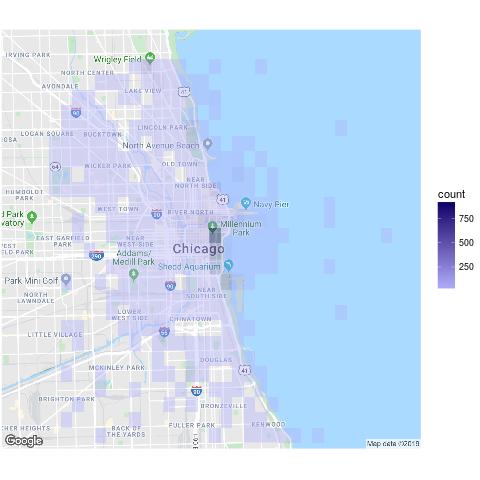}}
  \subfloat[Neutral]
            {\includegraphics[width=.45\textwidth]{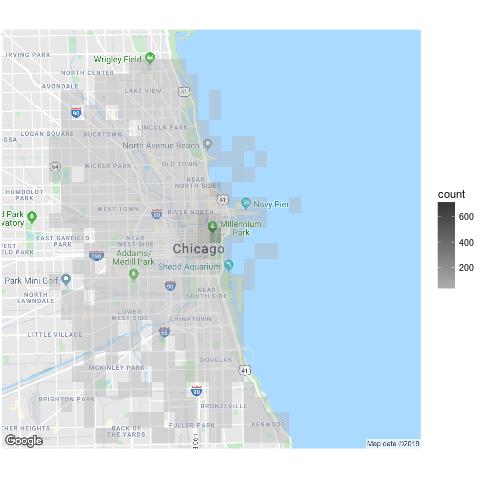}}\label{heatNeutral}
\subfloat[Negative]
            {\includegraphics[width=.45\textwidth]{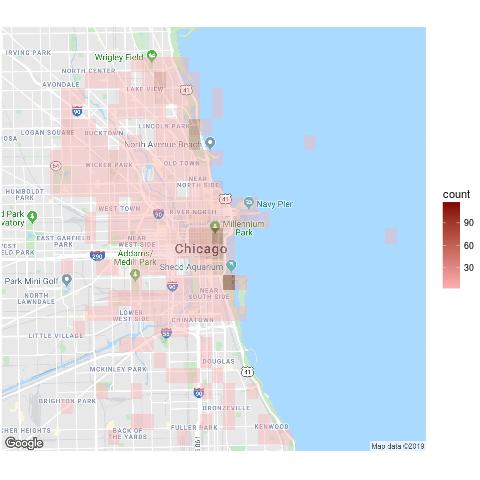}}\label{heatNegative}
\caption{Heatmap of photos according to their geolocation and sentiment.}
\label{heatmapImages}
\end{figure}

To evaluate the results of the proposed approach in a real-world scenario, we select an image subset from a publicly available dataset of media objects (images and videos) posted on Flickr 
We first extracted only geocoded images posted from Chicago, United States, obtaining 53,550 images. Next, we filter out images where the geolocation matches any building interior of the city. For this task, we explored building footprints of Chicago, which are publicly available in Chicago’s official open data portal\footnote{Chicago Open Data – https://data.cityofchicago.org.}. After this filter, we ended up with 17,469 images in the dataset. 
When applying the ConvNet of You~\etal~\cite{you2015robust} with YOLO + SUN attributes to infer the associated sentiment (our preferable approach due to its simplicity and comparable results with other approaches), we obtained 8,492 positive, 8,584 neutral, and 393 negative images. Figure \ref{heatmapImages} presents the heatmaps for each class (positive, neutral, and negative) according to image geolocation. 

Observe that downtown has a bigger concentration of all sentiment polarities, as expected. However, the higher concentration part in this area is a bit different between all the classes, especially for the negative one. In addition, it is also interesting to note a more prominent occurrence of positive images in the water, a rare phenomenon for negative photos. These differences are an indication that image sentiment could help to understand the subjective characteristics of different areas of the city, as shown by text content \cite{santosWI2018}.

\begin{figure}[!htb]
\includegraphics[width=.96\textwidth]{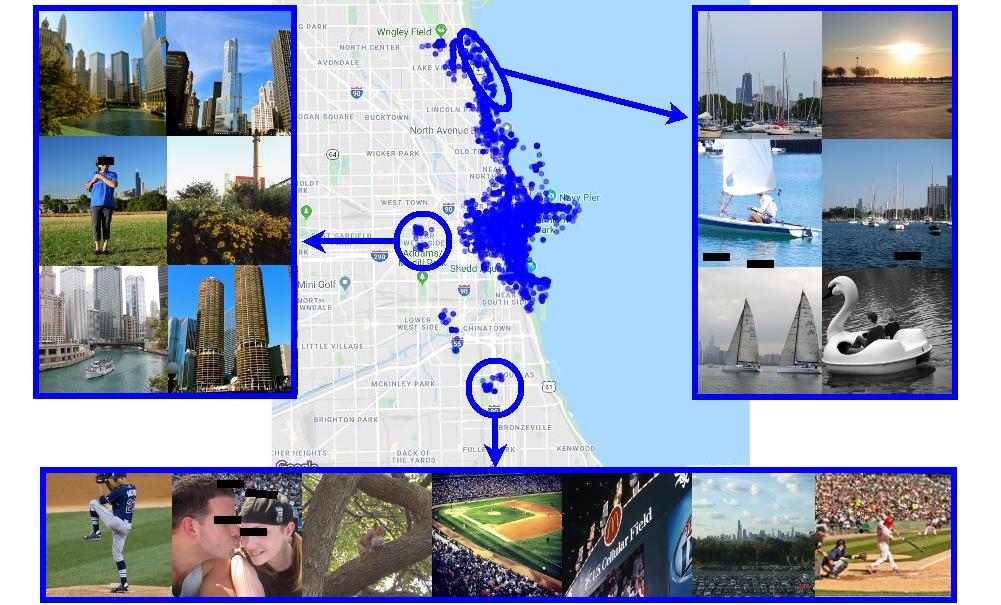}
\caption{Clusters found in Chicago for positive images and representative examples of different clusters.}
 \label{fig:clusterPositivo}
\end{figure}

\begin{figure}[!htb]
\includegraphics[width=.96\textwidth]{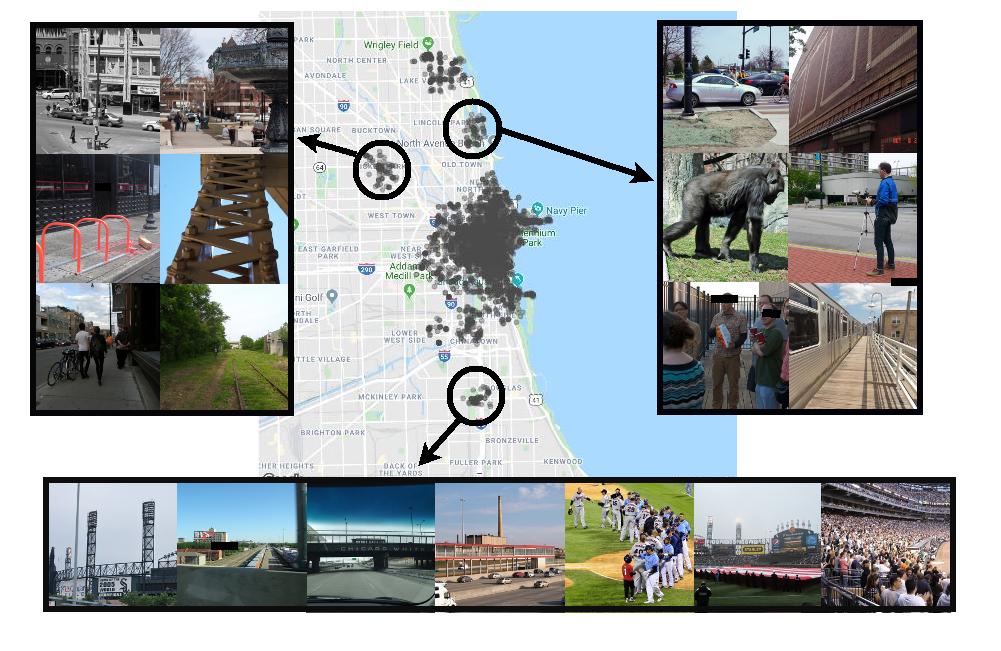}
\caption{Clusters found in Chicago for neutral images and representative examples of different clusters.}
 \label{fig:clusterNeutral}
\end{figure}

\begin{figure}[!htb]
\includegraphics[width=.96\textwidth]{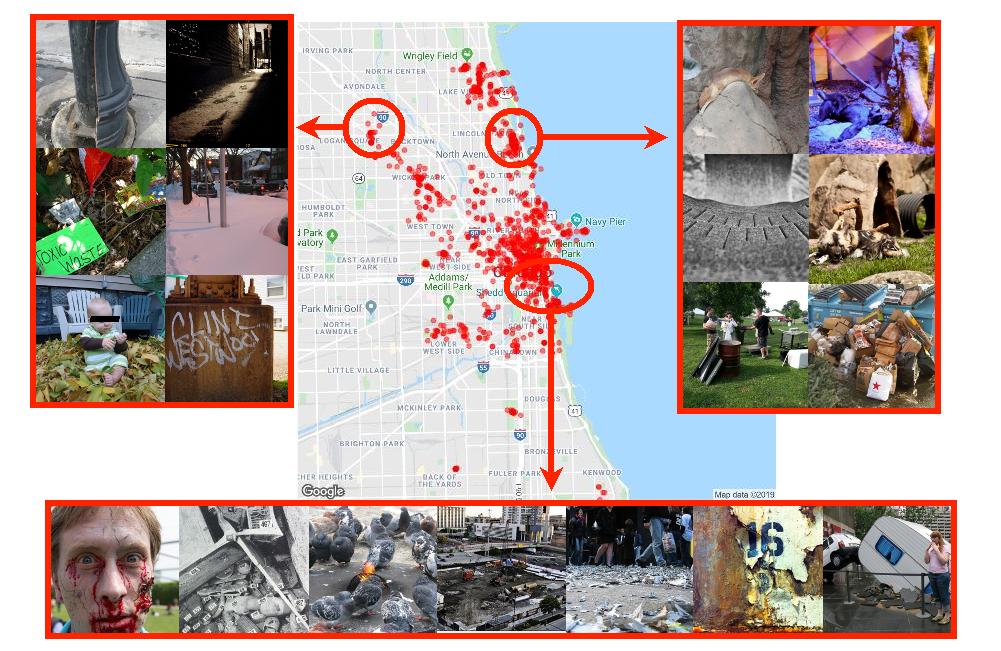}
\caption{Clusters found in Chicago for negative images and representative examples of different clusters.}
 \label{fig:clusterNegativo}
 \end{figure}

In order to favor this sort of evaluation, we performed a density-based clustering process using DBSCAN \cite{ester1996density}, an algorithm to find high-density regions separated from one another by regions of low-density. Points, images in our case, in low-density regions are classified as noise and ignored. DBSCAN requires two parameters: the minimum number of points needed to form a dense region $minPts$; and the radius $eps$. We did this clustering process for positive, neutral, and negative images. For the positive and neutral images we set the $eps$ parameter equal to $0.0045$ and considered $minPts=50$. For negative images, the $eps$ and $minPts$ parameters were set to $0.005$, and $10$, respectively. $Eps$ values were chosen following the approach proposed by Ester~\etal~\cite{ester1996density}.
Figures \ref{fig:clusterPositivo}, \ref{fig:clusterNeutral}, \ref{fig:clusterNegativo} show the clustering results for positive, neutral, and negative images, respectively, thus highlighting areas according to this feature. Apart from downtown and some parts of the coast, we do not see much agreement between negative clusters and positive (and neutral) ones. 

The performance of the positive and neutral classification, in general, is good. However, one could argue that a few positive images should be neutral and vice versa. For instance, most people could classify the baseball stadium at the bottom of Figure~\ref{fig:clusterNeutral} as positive. Those cases are expected to happen due to the subjective nature of the problem. 

Analyzing the negative images reported by our classifier (Figure \ref{fig:clusterNegativo}), we can find more significant mistakes. For instance, a baby smiling, a family making a picnic in the park, and some animals resting in the zoo. 

Despite these problems, we believe that the performance is acceptable for a wide range of tasks because most of the instances were correctly classified as negative. A possible way to minimize such errors would be to perform aggregated analysis, for example, consider a group of nearby images in the same area. The design of new semantic features, explicitly oriented for the sentiment analysis of outdoor urban scenes, could also help in this context.

To investigate the association of the classified images with some demographic characteristics of geographic areas, we also performed an analysis considering the median income of households per census tract. We got the publicly available data provided in the American Community Survey, administered by the United States Census Bureau\footnote{https://www.census.gov/programs-surveys/acs.}. We considered the study of 2015 because it covers the most recent period of the Flickr dataset studied.

We then mapped each image in one census tract, grouping each class of image in:  Low Income, where the median household income $i$ per year is less than $\$50k$; Medium Income, with $50 \leq i \leq 100$; and High Income where the median household income is bigger than $\$100k$.

\begin{figure}[!htb]
\includegraphics[width=.99\textwidth]{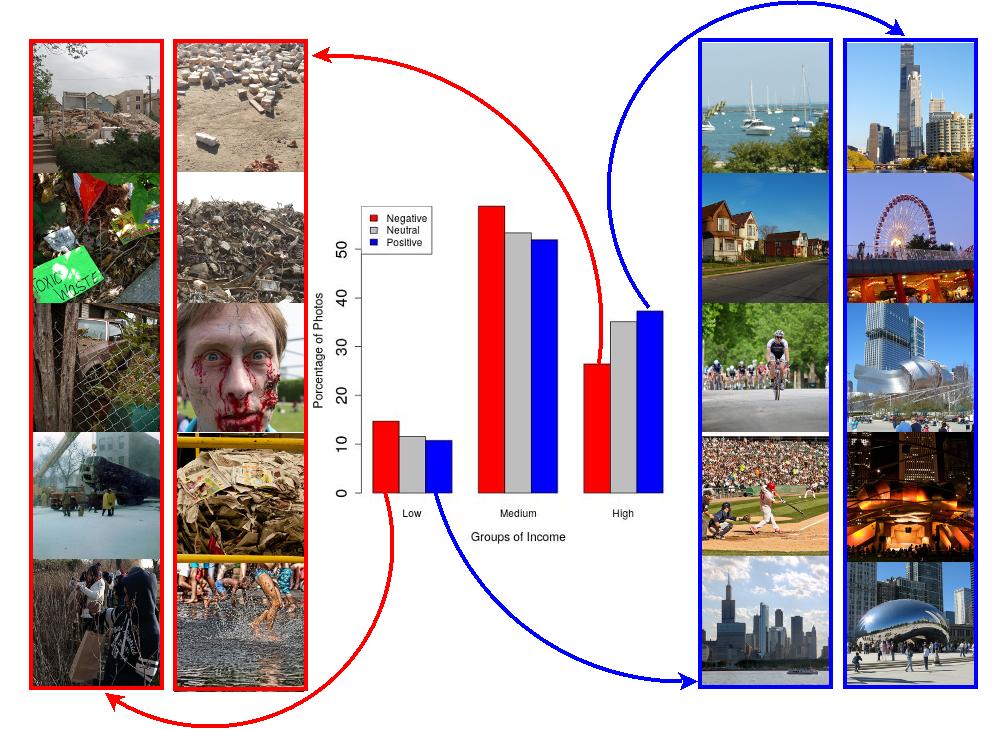}
\caption{Percentage of images according to different levels of income, considering positive, neutral, and negative images. The ranges are: low income (median household income $i$ per year $i< \$50k$), medium income ( $\$50k \leq i \leq \$100k$), and high income ($i>\$100k$). For low income and high income we show some representative images (indicated by red and blue arrows, respectively). }
 \label{figBarIncome}
\end{figure}

Figure \ref{figBarIncome} presents the results, as well as some representative images (indicated by arrows) for low and high income. In general, in the low income areas, we have fewer images of all sentiment classes and a slightly smaller number of positive images. 

This tendency is also observed for the medium income areas. However, when looking at the high income areas, this tendency is inverse, and we have a significantly bigger number of positive images. This result is interesting because it suggests that different classes of images correlate with certain areas of the city.

Investigating the images of the low and high income groups, we find that the overall classification is coherent. As illustrated in Figure \ref{figBarIncome}, most of the negative and positive images in all groups tend to reflect correctly the sentiment associated, especially in the positive ones. Again, we find that more probably mistakes are observed in the negative class, especially in the high income group. One could argue, for example, that the three bottom examples for the negative class of the high income group might be incorrectly classified. The same is valid for the bottom image of the examples for the negative class of the low income group. 

We also observe some differences between groups regarding the same sentiment. For instance, positive images in the low income group tend to reflect more outdoor sports, nature, and houses. Nevertheless, for the high income group, we tend to observe more images related to sculptures, buildings, and sights. These aspects can be further explored in other works, where more information about the groups is available.
\section{Discussion}\label{secLimitacoes}

Recent studies have shown that some information obtained from social media has the power to change our perceived physical limits as well as to help better understand the dynamics of cities. In this direction, there are some efforts to classify city areas under different aspects, for example, regarding the smell, cultural habits, noise, and visual aspects \cite{maisonneuve2009noisetube, quercia2014shortest, quercia2015smelly,le2015soho,SILVA201795,Mueller2017}.

This classification of urban areas may be useful for a variety of new applications and services. An example would be a new route suggestion tool that suggests the most visually pleasing way, which might be interesting for users in leisure time in the city, as was discussed in~\cite{quercia2014shortest}. Our study has the potential to complement these proposals, considering another aspect in this direction: the sentiment about urban outdoor environment.

In addition, the information that can be obtained automatically from our work can help fields of study where the collection of similar information occurs in ways that do not scale easily, such as interviews and questionnaires.  From the categorization of outdoor areas of the city according to the sentiment opinion understood by users, new socioeconomic studies on a large scale can be developed. By correlating these results with demographic indicators such as income and occurrence of crimes, non-obvious patterns can be understood, which may be useful in better urban planning and strategic public policies.

In this sense, it is worth mentioning the theory of ``broken windows'' proposed by Kelling \& Coles. The idea behind this theory is that the appearance of outdoor areas can impact on neighborhood safety reality: a broken window leads to another and, in turn, to future crimes \cite{kelling1997fixing}. This and other theories can be revisited on a large scale by exploring our study.

Our study has some possible limitations. One is concerning the labeling of the OutdoorSent dataset. Although we have counted on the collaboration of a diverse group of volunteers, we do not have the opinion of all the population strata. This means that the labels can be biased to specific groups of people. 
Another limiting factor of our proposal is the performance of our classifier. This is associated with the type of chosen images that becomes more challenging because of the greater diversity of possible variations. 

It is noteworthy that we improved the state-of-the-art. Still, we believe it may be possible to improve the classification performance with other approaches and techniques in future efforts.

As another contribution, we conducted a detailed analysis to understand the main elements shared by users on social networks for outdoor scenes and their connection with sentiment polarity. 

\begin{figure}[!htb]
\includegraphics[width=1\textwidth]{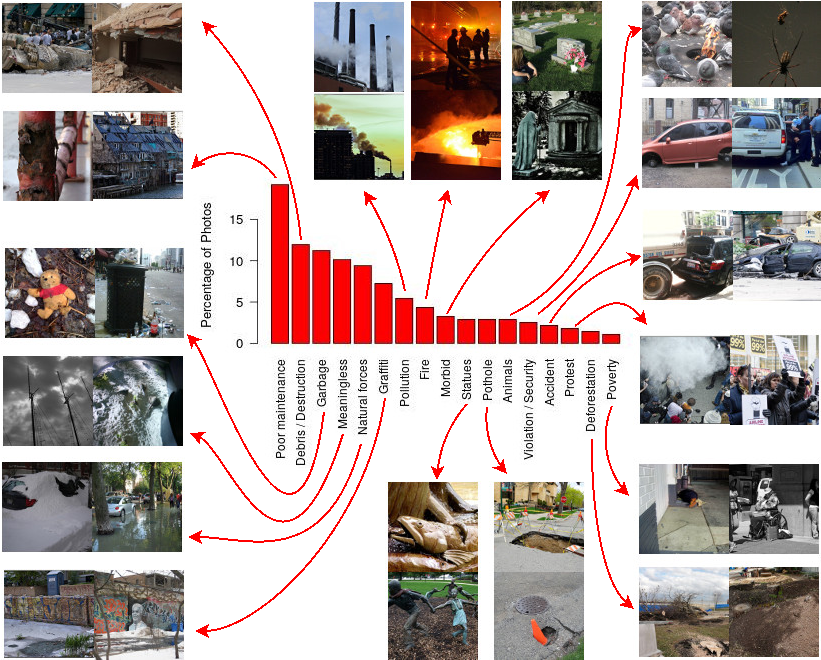}
\caption{Percentage of photos in each negative class with examples.}
 \label{fig:negativeExamples}
\end{figure}

\begin{figure}[!htb]
\includegraphics[width=.95\textwidth]{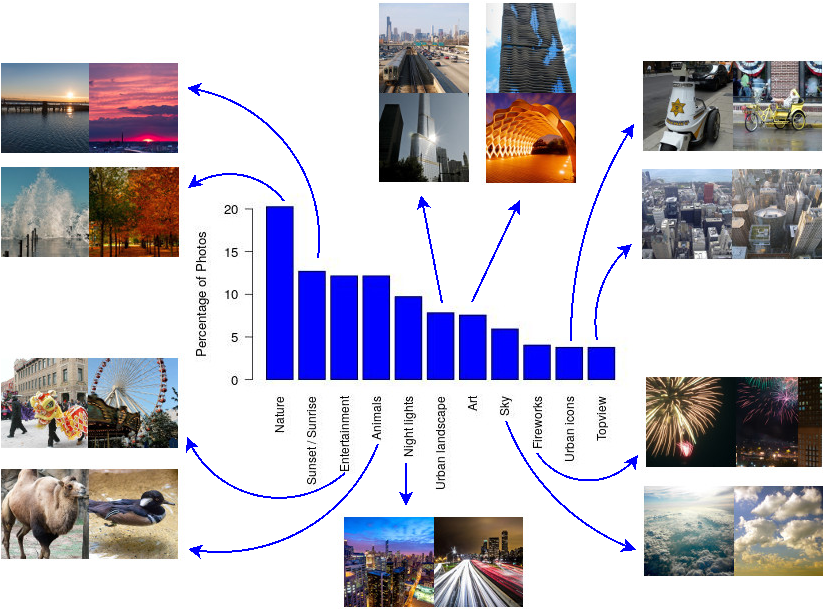}
\caption{Percentage of photos in each positive class with examples.}
 \label{fig:postivieExamples}
\end{figure}

From this study, using the proposed OutdoorSent dataset, we noticed that the majority part of negative images posted by the Flickr users could be mapped in a few classes. Most of them ($\approx$ 80\%) are related to common problems of big cities, such as poor maintenance, debris from construction sites, garbage, disturbances due to natural forces, graffiti, pollution, fire, and potholes (see Figure~\ref{fig:negativeExamples} for some examples of each class). On the other hand, images classified as positive can be mapped in an even smaller number of classes. We observe that most of them ($\approx$ 83\% of total images) are related to nature views, sunset pictures, entertainment activities, animals, panoramic views of the city during the night, urban landscapes, and art monuments, see Figure~\ref{fig:postivieExamples}.

\begin{figure}[!htb]
\centering
 \subfloat[]{\includegraphics[width=.4\textwidth]{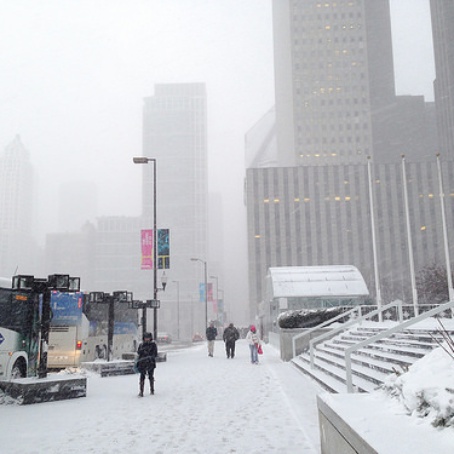}} \hspace{15pt}
  \subfloat[]{\includegraphics[width=.4\textwidth]{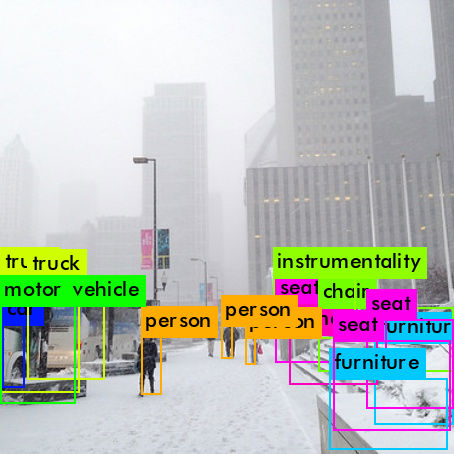}}
\caption{YOLO9000 detection/classification results: although important semantic attributes were found for the outdoor scene in (a), there are also many misclassifications such as chair, furniture and instrumentality (b). The design of a novel YOLO network, more focused of urban scene elements related to sentiment analysis could be a promising direction for future research on this field.}
 \label{fig:yolo_broad}
\end{figure}

Such understanding may have important implications for the design of new semantic features, explicitly oriented for the sentiment analysis of outdoor urban scenes. For instance, we noted that the problem domain addressed by YOLO9000~\cite{you2015robust} is too broad, being many categories related to indoor scenes, leading to many false detections and misclassifications, see Figure~\ref{fig:yolo_broad}. On the other hand, important categories related to sentiment analysis in outdoor scenes are missing, such as pollution, debris/rubble, and poor/bad maintenance. 

The same conclusions hold for the SUN~\cite{SUN1,SUN2} network, were some scene attributes do not seem to have a strong connection with our problem domain, such as carpet, vinyl, and gaming. Although the improvement of such features is beyond the scope of this work, we believe that the design of semantic concepts more focused on a cohesive set of urban scene elements is a promising direction for future research on this field and can be explored in further research works on image-based sentiment analysis.

\section{Conclusion}\label{secConclusion}

This study investigates if semantic attributes (YOLO and SUN) help to enhance the performance of ConvNets for sentiment analysis in outdoor images. Our experiments considered five different ConvNets, four frequently used in the machine learning area, and one designed for sentiment analysis in images.  We examined different datasets in the experiments, one of them proposed in this study (OutdoorSent). We find that semantic features improved the performance results of previous initiatives. Furthermore, we also show that
the use of semantic attributes improved the performance of all ConvNets architectures but had a much more significant impact on the most straightforward architectures.
Besides, we studied the impact on classification results of considering indoor images in the learning phase, not observing significative gain that justifies the higher cost demanded. We also performed a  cross-dataset generalization investigation. Surprisingly, the simplest ConvNet, was the most robust in this regard in all considered scenarios. In this direction, we found an indication that in this task, image editions, such as filters, and also the dataset size, plays an important role. In general, the results of this cross-dataset generalization experiment are promising, given the diversity of such datasets and the low average error.  We also showed in a real-world scenario that our results can help to understand the subjective characteristics of different areas of the city, helping to leverage new services. As future studies, there are several directions. In particular, by analyzing the OutdoorSent dataset carefully, we observed that the most negative and positive images could be mapped in a few classes. This indicates an opportunity of extending YOLO/SUN-like architectures with more specific classes for the outdoor context. This might also be the case for other contexts. 


\section*{Acknowledgements} 
This study was financed in part by the Coordenação de Aperfeiçoamento de Pessoal de Nível Superior - Brasil (CAPES) - Finance Code 001. This work is also partially supported by the project URBCOMP (Grant \#403260 /2016-7 from CNPq agency) and GoodWeb (Grant \#2018/23011-1 from São Paulo Research Foundation - FAPESP). We gratefully acknowledge the support of NVIDIA Corporation with the donation of the Titan Xp GPU used for this research. The authors would like to thank also the research agency Fundação Araucária.

\bibliographystyle{ACM-Reference-Format}
\bibliography{main}

\end{document}